\documentclass[journal]{IEEEtran}

\usepackage{cite}
\usepackage[hidelinks]{hyperref}
\usepackage{amsmath,amssymb,amsfonts}
\usepackage{mathtools, nccmath}
\DeclarePairedDelimiter{\nint}\lfloor\rceil

\usepackage{algorithm2e}
\usepackage{graphicx}
\usepackage{textcomp}
\usepackage{multirow}
\usepackage{multicol,lipsum,xparse}
\usepackage{booktabs, capt-of, tabularx}
\usepackage[dvipsnames]{xcolor}

\usepackage{tikz}
\usetikzlibrary{shapes.geometric, arrows}
\tikzstyle{startstop} = [rectangle, rounded corners, minimum width=3cm, minimum height=1cm,text centered, draw=black, fill=red!30]
\tikzstyle{io} = [trapezium, trapezium left angle=70, trapezium right angle=110, minimum width=3cm, minimum height=1cm, text centered, draw=black, fill=blue!30]
\tikzstyle{process} = [rectangle, minimum width=3cm, minimum height=1cm, text centered, draw=black, fill=orange!30]
\tikzstyle{decision} = [diamond, minimum width=3cm, minimum height=1cm, text centered, draw=black, fill=green!30]
\tikzstyle{arrow} = [thick,->,>=stealth]

\begin{document}

\title{CIMulator: A Comprehensive Simulation Platform for Computing-In-Memory Circuit Macros with Low Bit-Width and Real Memory Materials}


\author{Hoang-Hiep Le, Md. Aftab Baig,  Wei-Chen Hong, Cheng-Hsien Tsai, Cheng-Jui Yeh, Fu-Xiang Liang, I-Ting Huang, Wei-Tzu Tsai, Ting-Yin Cheng, Sourav De, Nan-Yow Chen, Wen-Jay Lee, Ing-Chao Lin, Da-Wei Chang, Darsen D. Lu$^*$

*e-mail: darsenlu@mail.ncku.edu.tw}



\maketitle

\begin{abstract}
This paper presents a simulation platform, namely CIMulator, for quantifying the efficacy of various synaptic devices in neuromorphic accelerators for different neural network architectures. Nonvolatile memory devices, such as resistive random-access memory, ferroelectric field-effect transistor, and volatile static random-access memory devices, can be selected as synaptic devices. A multilayer perceptron and convolutional neural networks (CNNs), such as LeNet-5, VGG-16, and a custom CNN named C4W-1, are simulated to evaluate the effects of these synaptic devices on the training and inference outcomes. 
The dataset used in the simulations are MNIST, CIFAR-10, and a white blood cell dataset. By applying batch normalization and appropriate optimizers in the training phase, neuromorphic systems with very low-bit-width or binary weights could achieve high pattern recognition rates that approach software-based CNN accuracy. We also introduce spiking neural networks with RRAM-based synaptic devices for the recognition of MNIST handwritten digits.
\end{abstract}


\IEEEpeerreviewmaketitle
\section{Introduction}

\IEEEPARstart {C}{omputation} for complex deep neural network (DNN) algorithms require substantial time and energy if performed using traditional computing systems due to the bottleneck of data transfer between memory and processing units; this is known as the von Neumann bottleneck \cite{von_Neumann_bottleneck_1, von_Neumann_bottleneck_2}. Numerous hardware \cite{hardware_Kala, hardware_Ghanbari} and software solutions for accelerating computations while maintaining accuracy have been proposed for big data applications, such as computer vision and speech and natural language recognition; however, such solutions still have potential for further improvement. For example, the use of two powerful graphical processing units (GPUs) to train AlexNet to categorize images on the ImageNet database requires at least 5 days \cite{ImageNet_GPU}. Existing CPU-based von Neumann hardware architectures have two key shortcomings: First, CPU-centric systems have only a handful of computational cores and therefore a handful of threads running in parallel. Even given a high clock frequency, data throughput is still limited. This can be overcome by using hardware with numerous processing elements (PEs), such as GPUs, artificial intelligence (AI) accelerators (e.g., Eyeriss \cite{Eyeriss}) or tensor processing unit (TPU) \cite{TPU}. Second, computation (CPU or GPU) and memory units are separated in conventional systems. Moving data between computation and memory units requires substantial time and power, thus creating a performance and power consumption bottleneck.

Computing-in-memory (CIM), a type of neuromorphic computing, is a
highly efficient hardware solution that could address these challenges. CIM
combines computation and memory in the same circuit; it is a parallel solution
in which a single memory device performs both data storage and multiply-accumulate (MAC) operations. CIM can drastically increase the energy efficiency of data-intensive computations compared with conventional digital
computing \cite{NE, NN, IEEE}. Moreover, CIM overcomes the von Neumann bottleneck
drawback by directly deploying MAC inside the accelerator without
transferring data between memory and processing units. Thus, some basic
mathematical problems, such as finding solutions for a system of linear
equations, can be solved in only one step \cite{PNAS} through on-site inverse matrix
multiplication in a CIM accelerator.

Benchmarking a CIM-based neuromorphic computation methodology is
crucial for determining its feasibility in real applications. Such benchmarking
involves simulating and evaluating the performance of neuromorphic accelerators
in terms of factors such as circuit design, recognition accuracy, energy consumption, and latency. The benchmarking results enable the selection
of appropriate synaptic devices and hardware architectures for the
target applications and dataset. A study \cite{PCM02} demonstrated an on-chip machine
learning system for large-scale artificial neural networks by using nonvolatile
memory (NVM)-based synapses that could be completely simulated
on a simulation platform. Other simulators such as PRIME \cite{PRIME}, MNSIM \cite{MNSIM}, and SIAM \cite{SIAM} have been demonstrated to flexibly enable system-level design and to be useful references for CIM system simulation platforms. However, these systems all have limitations. Specifically, the systems proposed
in \cite{PRIME} and \cite{MNSIM} have limited applicability for devices with nonideal properties, and the system proposed in \cite{SIAM}, which focuses on circuit architecture design optimization, is available for only static random-access memory (SRAM)- and resistive random-access memory (RRAM)-based synaptic devices. The NeuroSim platform \cite{NeuroSim1, NeuroSim2} mostly addresses these limitations; however, it has low online training accuracy for systems with less than 6-bit weights. To address the aforementioned limitations, this paper presents a simulation framework, namely CIMulator, that involves functional simulation and performance estimation procedures for neuromorphic accelerators. The platform facilitates the investigation of the characteristics of neuromorphic devices and the establishment of efficient training algorithms. The main contributions of this paper are summarized as follows.

\begin{enumerate}[] 
\item 
We present a simulation platform, CIMulator, for benchmarking CIM-based neuromorphic hardware with a variety of neural network architectures, synaptic devices, and datasets.
\item
We propose a novel algorithm for updating and storing weights in the
cells of 2D1S \cite{2D1S} crossbar arrays during training. It makes update process simpler and alleviates influence of asymmetric characteristic of synaptic devices.
\item
Our platform supports the simulation of retraining or single-inference
tasks through the use of only pretrained models and weights to mimic offline
training. Weights may be obtained from external sources, such as PyTorch
packages, or prior runs of the CIMulator online training simulation.
\item
We investigate low-bit-width CIM systems, which reduces both memory usage (chip area) and computational cost time \cite{SWALP}. Binary weighted systems may also be immune to the influence of nonlinearity and variations because only two states are considered for synaptic memory. Low-bit-width systems may intrinsically have a lower recognition accuracy. Nevertheless, we propose remedies including accumulated weight update \cite{accumulated} and batch normalization \cite{Batchnorm} techniques, and demonstrate using our platform that these techniques can indeed achieve high accuracy in image recognition tasks.
\item
We upgrade the previous version of this simulation platform \cite{CIMulator_v1, ICASI_Lu, Thesis_WeiChen} to include benchmarking with a variety of neural network architectures, including a multilayer perceptron (MLP), LeNet-5, VGG-16, and a newly proposed convolutional neural network (CNN) named C4W-1, for various types of synaptic devices implemented on MNIST \cite{MNIST}, Cifar-10 \cite{Cifar10}, and white blood cell (WBC) image datasets \cite{WBC}.
\item
We use the CIMulator platform to calibrate various volatile and nonvolatile
memory technologies. CIMulator results were obtained for synaptic
arrays comprising several types of memory devices, including RRAM, ferroelectric
fin field-effect transistor (Fe-FinFET), and SRAM devices, that we
fabricated and measured.
\item
We add spiking neural networks (SNNs), also known as third generation neural networks, along with available MLP and CNNs as network architectures in the CIMulator platform. SNNs more closely mimic biological neural networks than conventional DNNs do; we believe that including SNNs can enable the execution of neuromorphic simulations on the platform.

\end{enumerate}

The rest of this article is organized as follows. Section II provides a detailed description of the CIMulator platform. Then, section III discusses various
CIM device types, namely RRAM-, Fe-FinFET-, and SRAM-based synaptic memory devices. Next, section IV presents various CIM architectures and simulation
results. Finally, Section V presents the conclusion.

\section{CIMulator - Simulation Platform for Neuromorphic Accelerator}

\subsection{CIM Hardware Architecture Assumptions}
\begin{figure}[ht]
\centerline{\includegraphics[width=1.0\linewidth]{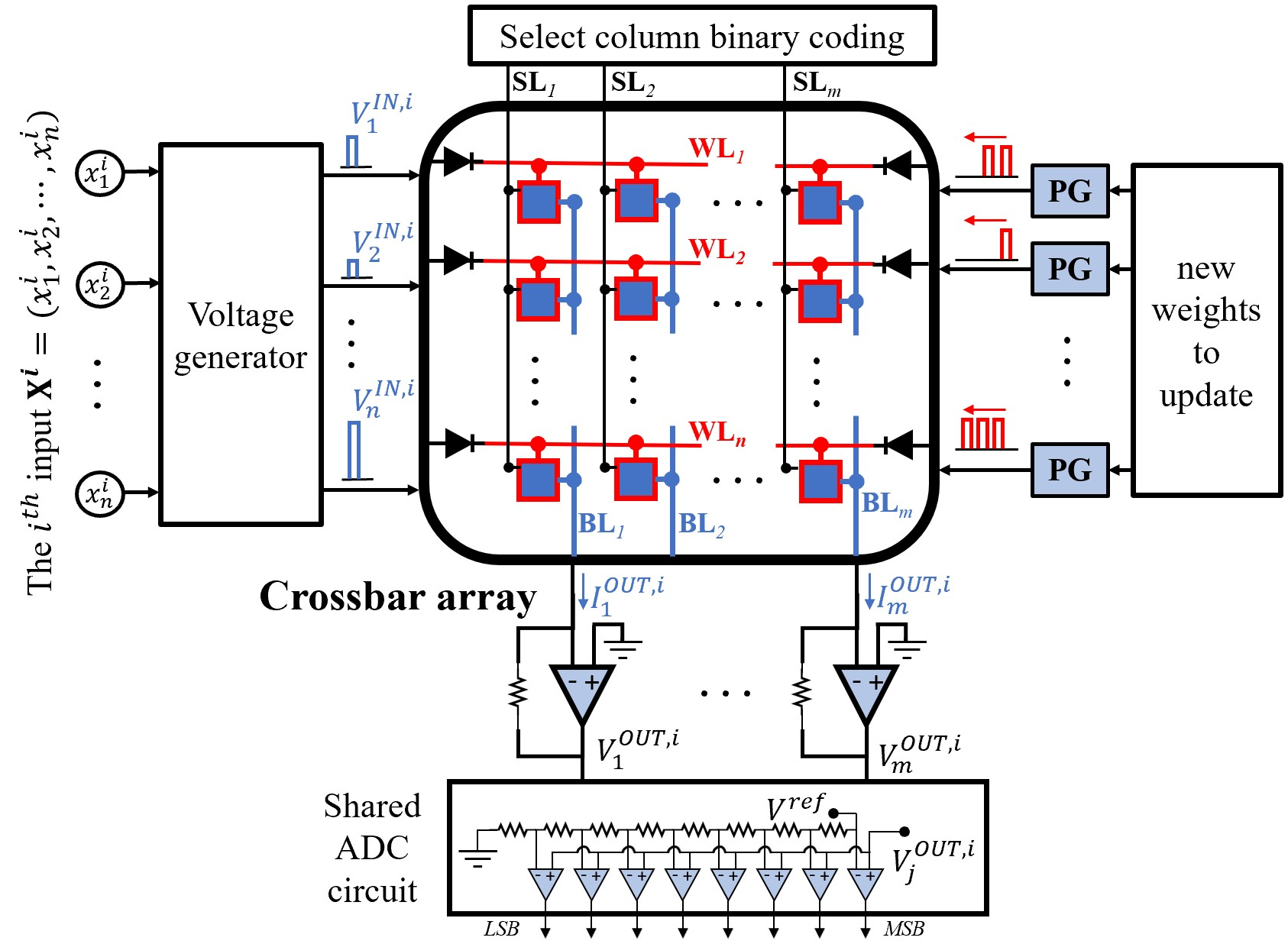}}
\caption{An illustration for a CIM circuit architecture}
\label{CIM_architecture}
\end{figure}

The fundamental CIM circuit architecture comprises a crossbar array and its peripheral circuitry is illustrated in Fig.~\ref{CIM_architecture}. The key components of the architecture are as follows: input, crossbar array, weight update, and select column.


First, the input block is a terminal gate that receives raw input data and transforms them into electrical signals for subsequent processing. In CIMulator, input data can be represented using binary or unary coding. The voltage generator block generates pulses with amplitudes proportional to the coded values. 
Second, the crossbar array contains synaptic cells that convert
the input voltage into current and collectively implement the MAC operation. The source line (SL) connects all of the devices to a binary coding circuit that selects one column at a time for performing the MAC operation. The word line (WL) connected to the bit line (BL) through synaptic cells is the gate for the input data. The current sum flows out along each column representing a
weighted sum at a neuron; this current is then converted into a voltage signal by an operational amplifier. A shared analog-to-digital converter (ADC) circuit in turn quantizes the outputs and transfers the obtained binary signals
to neurons in the next layer. Third, the weight update block computes
new weights and uses pulse generation to transform them into a series of
pulses. The generated pulses are used to program new weight values in the
synaptic cell write process. Three pulse generation schemes are described
in \cite{FeRAM}; however, only the first one, in which duration and amplitude of pulse are constant, is supported in the platform. Fourth, the column selection block is to guarantee the MAC operation, and also the weight updating in later steps, is executed on a single column of the crossbar array at once. The output value of the block is always a one-hot vector. Because the input and weight update blocks directly connect synaptic cells through common WLs, diodes are used in the rows of the array to avoid interference between the two processes. 

The MAC computation is the core operation of a layer-to-layer connection
in a neural network. For example, in a fully connected neural network, an
output layer (excluding the activation function) may be expressed as

\begin{equation}
\mathbf{y}= \mathbf{W} \mathbf{x} + \mathbf{b}
\label{MAC}
\end{equation}
where $\mathbf{y} = [y_1, y_2,..., y_m]^{'}$ and $\mathbf{b} = [b_1, b_2,..., b_m]^{'}$ are column vectors containing the values and biases, respectively, of the neurons in the output layer; $\mathbf{W}$ is an $m \times n$ matrix of weights corresponding to each input-to-output connection; and $\mathbf{x}= [x_1, x_2,..., x_n]^{'}$ is the column vector representing neuron values in the input layer (after the activation function). The matrix multiplication $\mathbf{W} \mathbf{x}$ in (\ref{MAC}) can be completely deployed in the accelerator as $\mathbf{G} V$, where $\mathbf{G}$ is the matrix of conductances of cells in the crossbar array (representing the weight matrix) and $V$ is the column vector of the input voltages (representing the input data). Clearly, the output vector $I=\mathbf{G}V=[I^{OUT}_1, I^{OUT}_2,..., I^{OUT}_m]$ represents the result of the matrix multiplication $\mathbf{W}\mathbf{x}$.

\begin{figure}[ht]
\centerline{\includegraphics[width=0.8\linewidth]{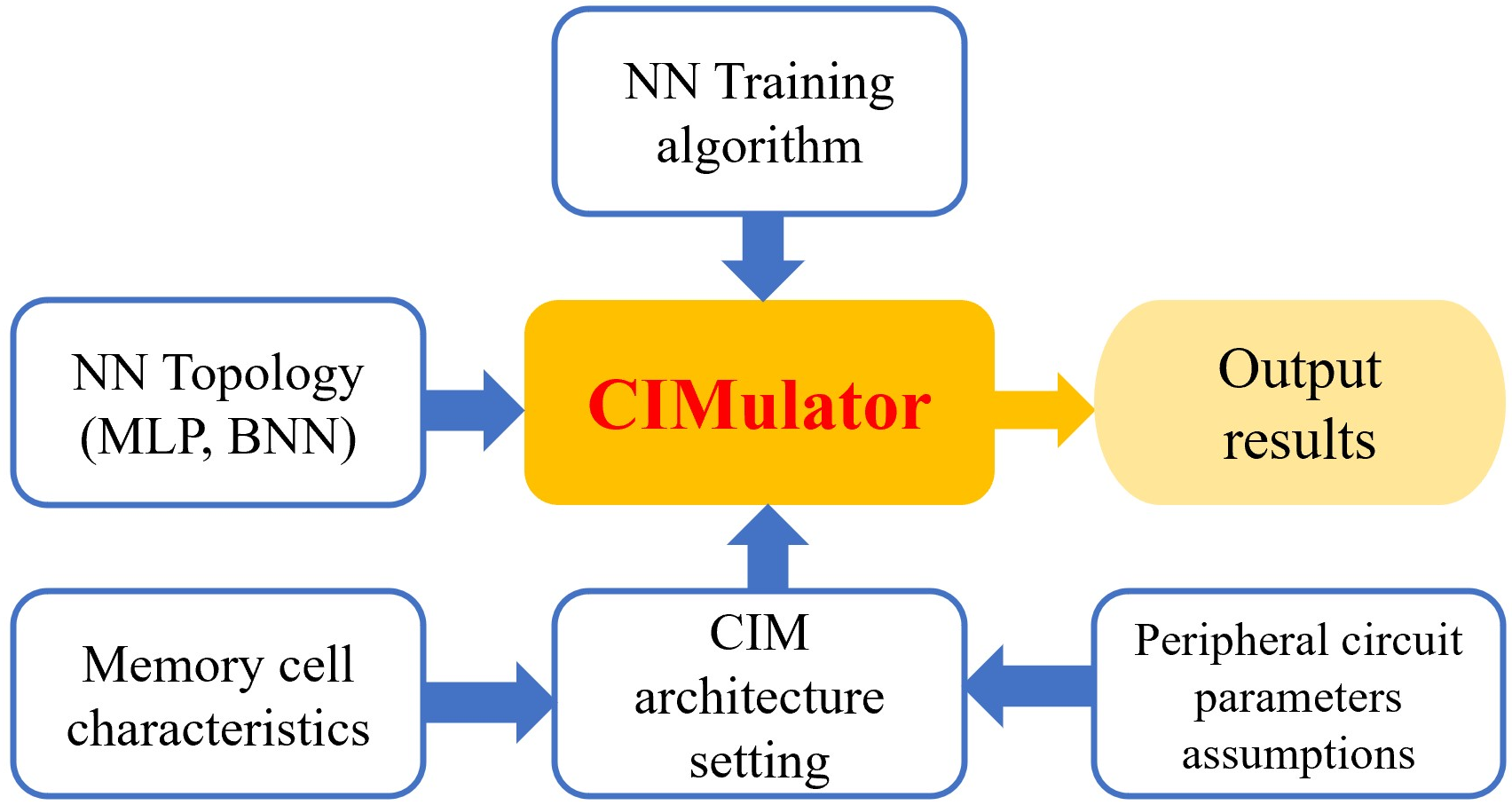}}
\caption{Neuromorphic simulation platform CIMulator architecture.}
\label{CIMulator_architecture}
\end{figure}

\begin{figure}[ht]
\centerline{\includegraphics[width=1.0\linewidth]{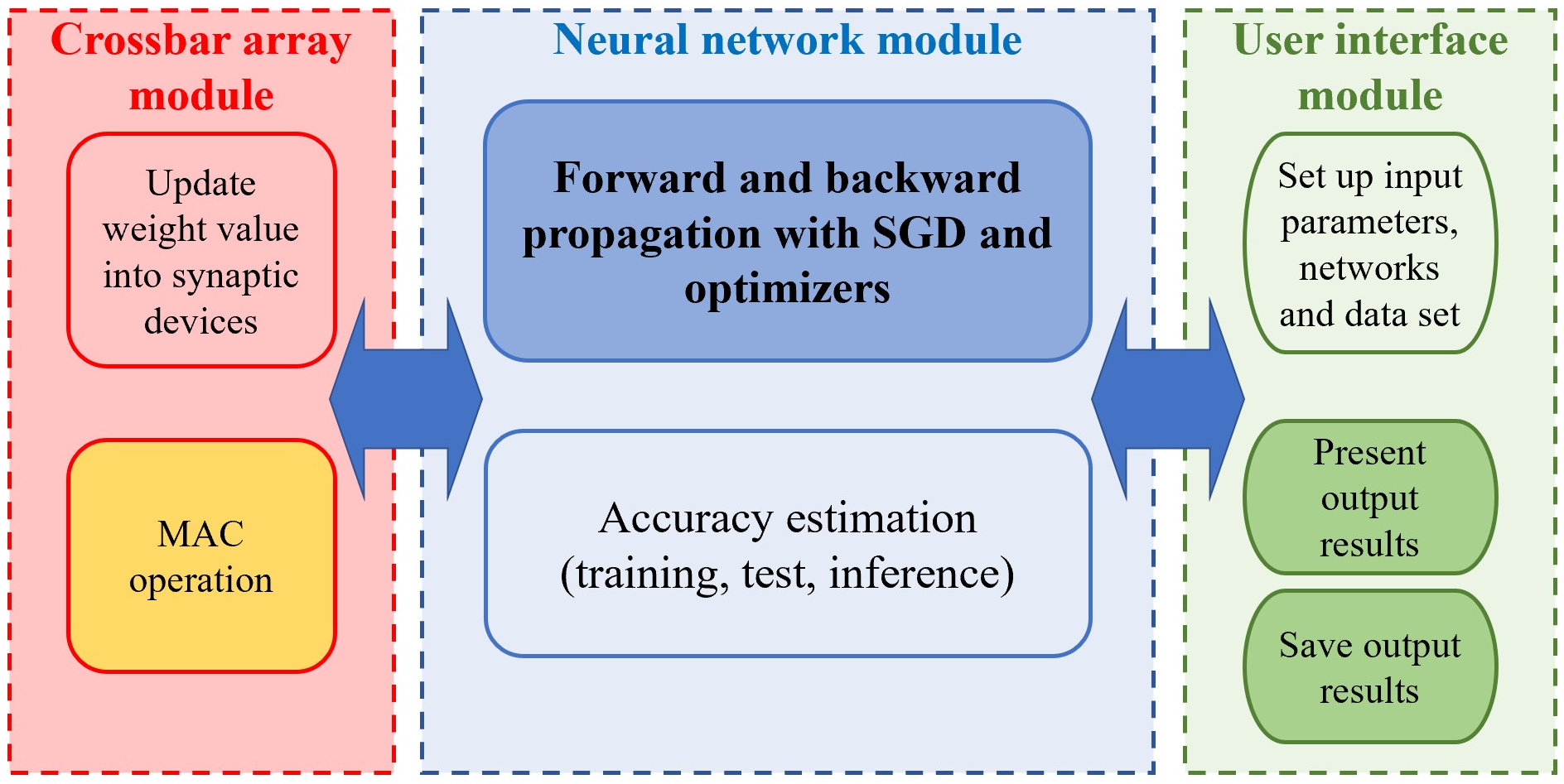}}
\caption{Processing modules of the CIMulator platform.}
\label{CIMulator_flowchart}
\end{figure}

The main goal of the CIMulator platform is to simulate such above-mentioned neuromorphic accelerators. Fig.~\ref{CIMulator_architecture} presents the key components of the software. Furthermore, the platform can be divided into three modules (user interface, neural network, and crossbar array), which are illustrated in Fig.~\ref{CIMulator_flowchart} and described as follows.



The user interface module is used to input the parameters of training dataset, synaptic device characteristics, cell design, crossbar array size, neural network architecture, training algorithm, peripheral circuit properties, and other settings. The neural network module implements algorithms for training and inference for a given neural network topology. In addition to applying the stochastic gradient descent (SGD) algorithm and backpropagation,  optimizers such as adaptive moment estimation \cite{Adam} and root mean squared propagation (RMSprop) \cite{RMSprop} can be optionally applied to increase accuracy and accelerate the training process. Finally, the crossbar array module simulates the CIM circuit and implements the WRITE and MAC operations. In detail, former is for weight updating, and latter is for computing weighted sum.

\subsection{Nonlinearity of Synaptic Devices}
\begin{figure}[ht]
\centerline{\includegraphics[width=0.8\linewidth]{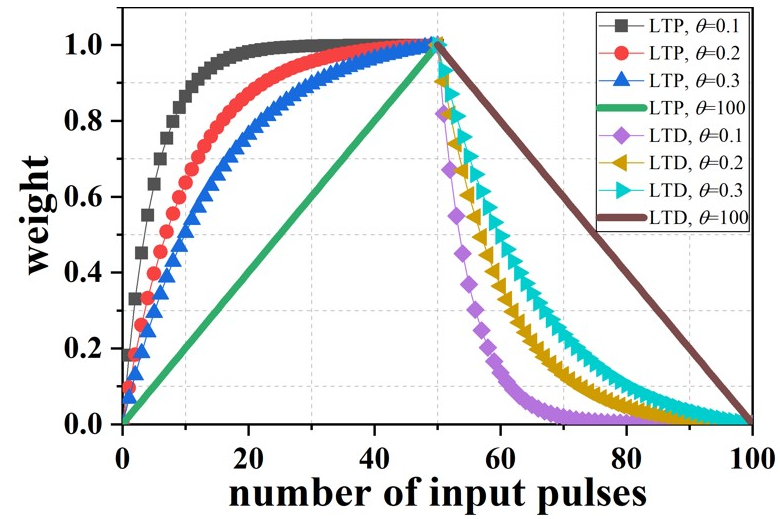}}
\caption{LTP and LTD curves with varying values of nonlinear ratio $\theta 's$}
\label{LTP_LTD}
\end{figure}

Nonlinearity \cite{Nonlinearity} is a key characteristic of nonideal synaptic devices and substantially affects the WRITE operation. Nonlinearity refers to the nonlinear relation between the number of input voltage pulses and the resulting conductance value of a set device. In the CIMulator platform, nonlinearity is represented by the ratio $\theta$; a smaller $\theta$ value indicates greater nonlinearity. Furthermore, nonlinearity can be characterized using LTP and long-term depression (LTD) curves. If these curves are asymmetric, $\theta_{LTP}\neq\theta_{LTD}$. Fig.~\ref{LTP_LTD} presents LTP and LTD curves for various  $\theta$ values. 

\subsection{Process Variation Modeling for Synaptic Devices}
\begin{figure}[ht]
\centerline{\includegraphics[width=0.8\linewidth]{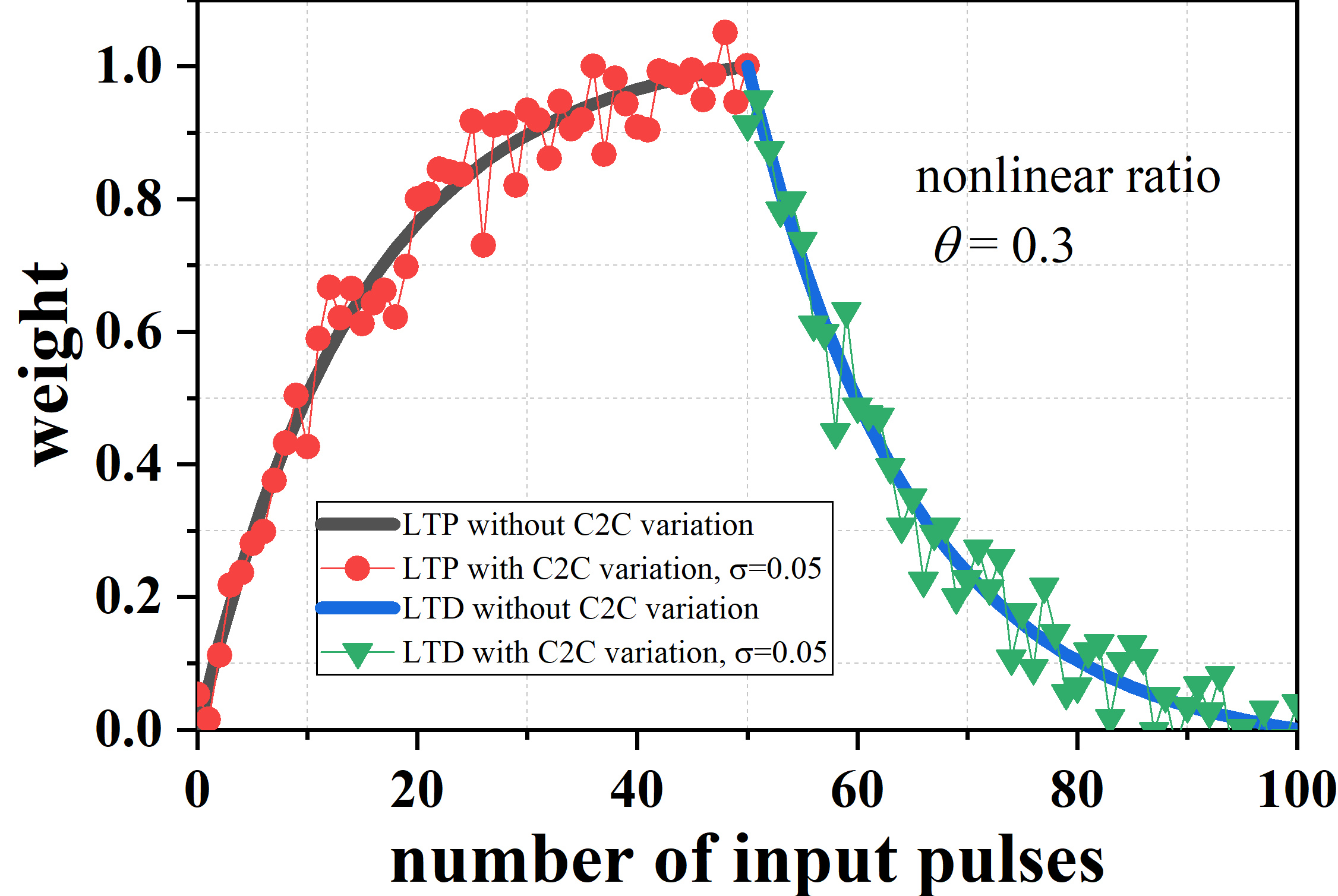}}
\caption{An illustration of C2C variation affecting value of weight}
\label{fig_C2C}
\end{figure}

In addition to nonlinearity, neuromorphic devices have inherent process induced variations that may substantially alter the outcome of the MAC operation, resulting in reduced accuracy during neural network training. CIMulator considers two types of random variation: cycle-to-cycle (C2C) and D2D. C2C is a variation after each READ or WRITE cycle, and D2D is a fixed discrepancy between devices on the same chip. In CIMulator, the value of a weight after the application of these variations is the sum of the programmed weight and the C2C and D2D variations. Both are defined by users and are normally distributed $\mathcal{N}(\mu,\sigma^2)$ or log-normally distributed $\mathcal{LN}(\mu_L,\sigma_L^2)$ random variables, where $\mu=0$, $\mu_L=0$ are means and $\sigma^2$, $\sigma_L^2$ are variances. These variations result in erroneous MAC values for forward inference and inaccurate weight updates during backward propagation. Fig.~\ref{fig_C2C} presents an example of the considerable shift in weight values given only C2C variation with $\sigma=0.05$ at $\theta$ = 0.3. In fact, D2D variation has small effect on the final accuracy if online training is employed because device variation can be compensated for during training weight adjustment. In contrast, this is not the case for C2C variation because random fluctuations occur in every cycle. A detailed investigation of this topic is available in \cite{Frontiers_De}.

\subsection{Single or Dual Device Synaptic Weight}
\begin{figure}[ht]
\centerline{\includegraphics[width=1.0\linewidth]{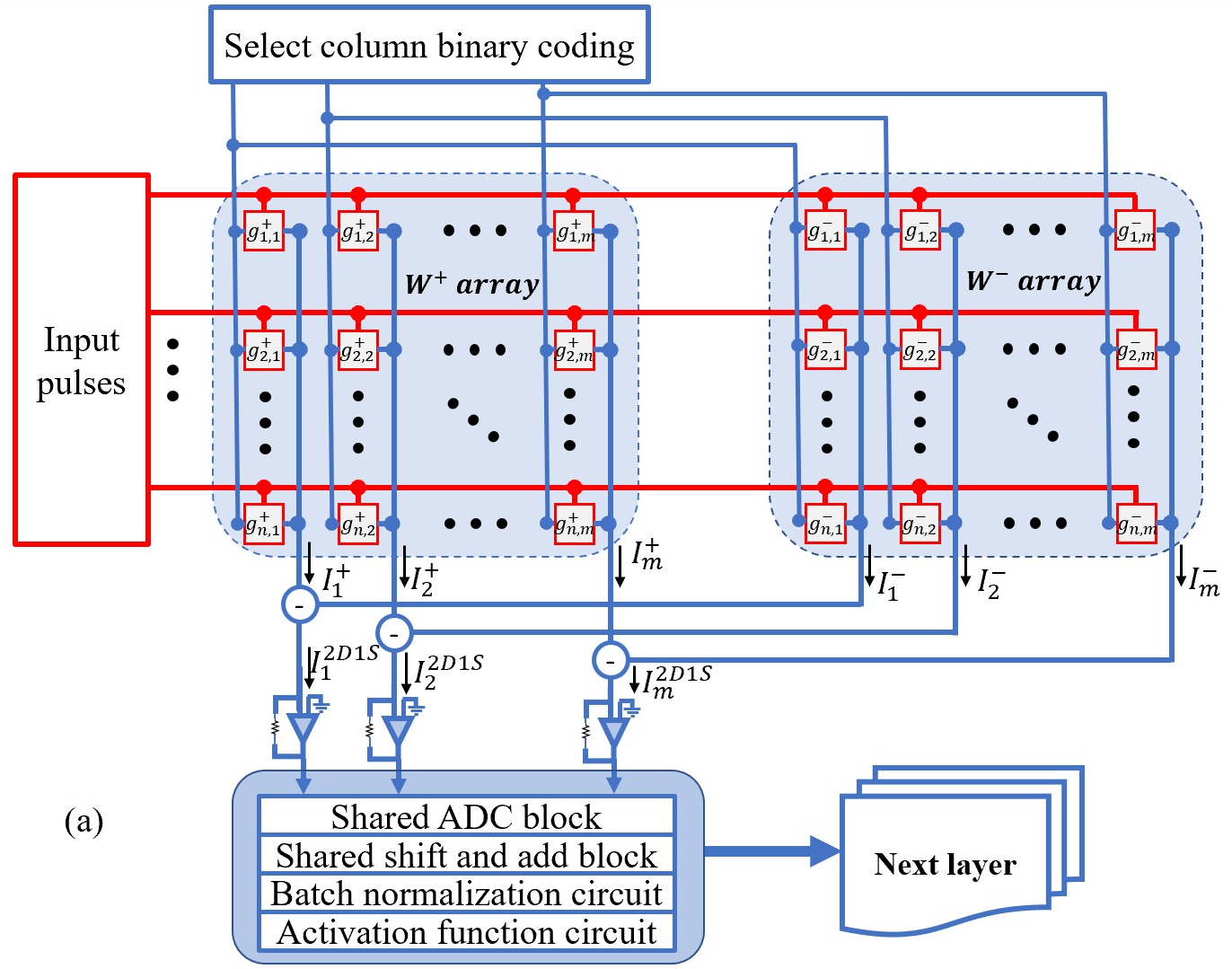}}
\par
\centerline{\includegraphics[width=0.9\linewidth]{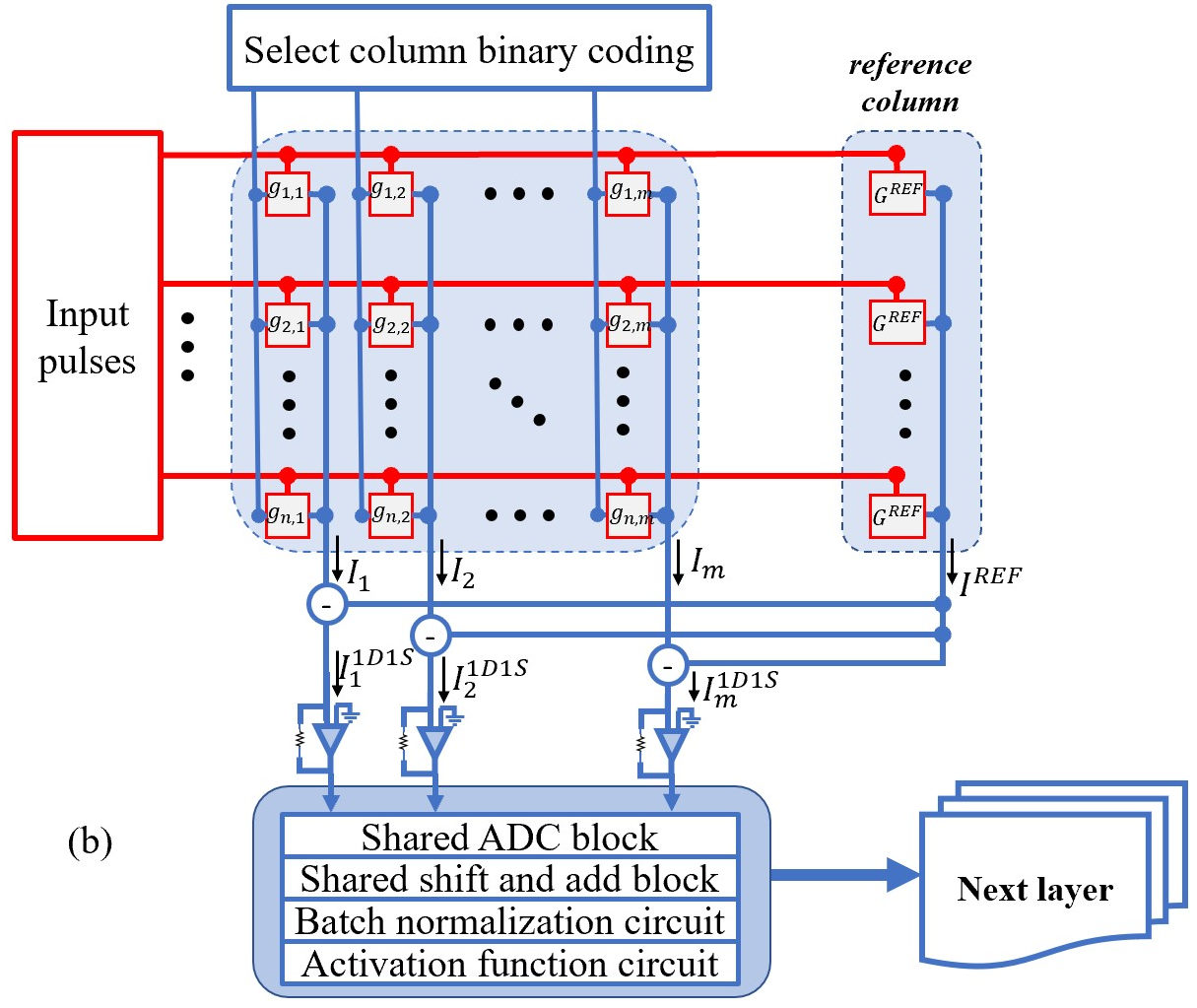}}
\caption{A conceptual diagram of (a) the 2D1S architecture circuit which includes two crossbar arrays for positive and negative weight, and (b) the 1D1S architecture which includes a single crossbar array and a reference column with constant conductance values.}
\label{1D1S_2D1S}
\end{figure}

Memory devices in synaptic arrays are connected in crossbar form, and their non-negative conductance must represent both positive and negative weight values. Two prominent array architectures have been proposed for solving this problem effectively. The two-device-one-synapse (2D1S) architecture\cite{2D1S} stores each weight in two memory devices [Fig.~\ref{1D1S_2D1S}(a)], whereas the one-device-one-synapse (1D1S) architecture \cite{1D1S_Agarwal, 1D1S_Truong} stores each weight in only one memory device [Fig.~\ref{1D1S_2D1S}(b)].

In the 2D1S architecture, an accelerator containing two separate crossbar arrays of positive and negative synaptic weights is used. The output currents are calculated as follows:

\begin{equation}
I^{2D1S}_i = I^+_i-I^-_i=\sum_{j=1}^{n}V_j(g^{+}_{j,i}-g^{-}_{j,i})
\label{eq1}
\end{equation}
where $I^+_i$, $I^-_i$ are the output currents of the $i^{th}$ columns of the positive and negative arrays, respectively; $V_j$ is the magnitude of the input voltage sum applied on the $j^{th}$ row; and $g^+_{j,i}$, $g^-_{j,i}$ are the conductances of the synaptic elements of the $j^{th}$ row and the $i^{th}$ column in the positive and negative arrays, respectively. Setting the conductance requires writing positive or negative pulses to the device. A single weight $w_{j,i}$ is represented as the difference in $g^{+}_{j,i}$ and $g^{-}_{j,i}$ between the synaptic devices in two separated crossbar arrays. Because multiple combinations of conductance values can represent a single weight, the actual method of conductance setting depends on the weight update rule. Consider $b$ be bit weight of a synaptic element, in non-binary weight system ($b>1$) the total number of levels of representable single weights should be $2^{b+1}-1$. Otherwise, in binary weight system ($b=1$), the platform would apply two-level weight representation scheme which guarantees values of single weight is $-1$ or $1$ only.

In the 1D1S architecture, the accelerator comprises only one crossbar array and a reference column with fixed conductance $G^{REF}$ set for all synaptic cells. Hence, the weight value is the difference in conductance between the main synaptic device and $G^{REF}$. This architecture eliminates the additional synaptic arrays in the 2D1S architecture. The output current can be expressed as follows:

\begin{equation}
I^{1D1S}_i = I_i-I_{REF}=\sum_{j=1}^{n}V_j(g_{j,i}-G^{REF})
\label{eq2}
\end{equation}

\begin{figure}[ht]
\centerline{\includegraphics[width=0.8\linewidth]{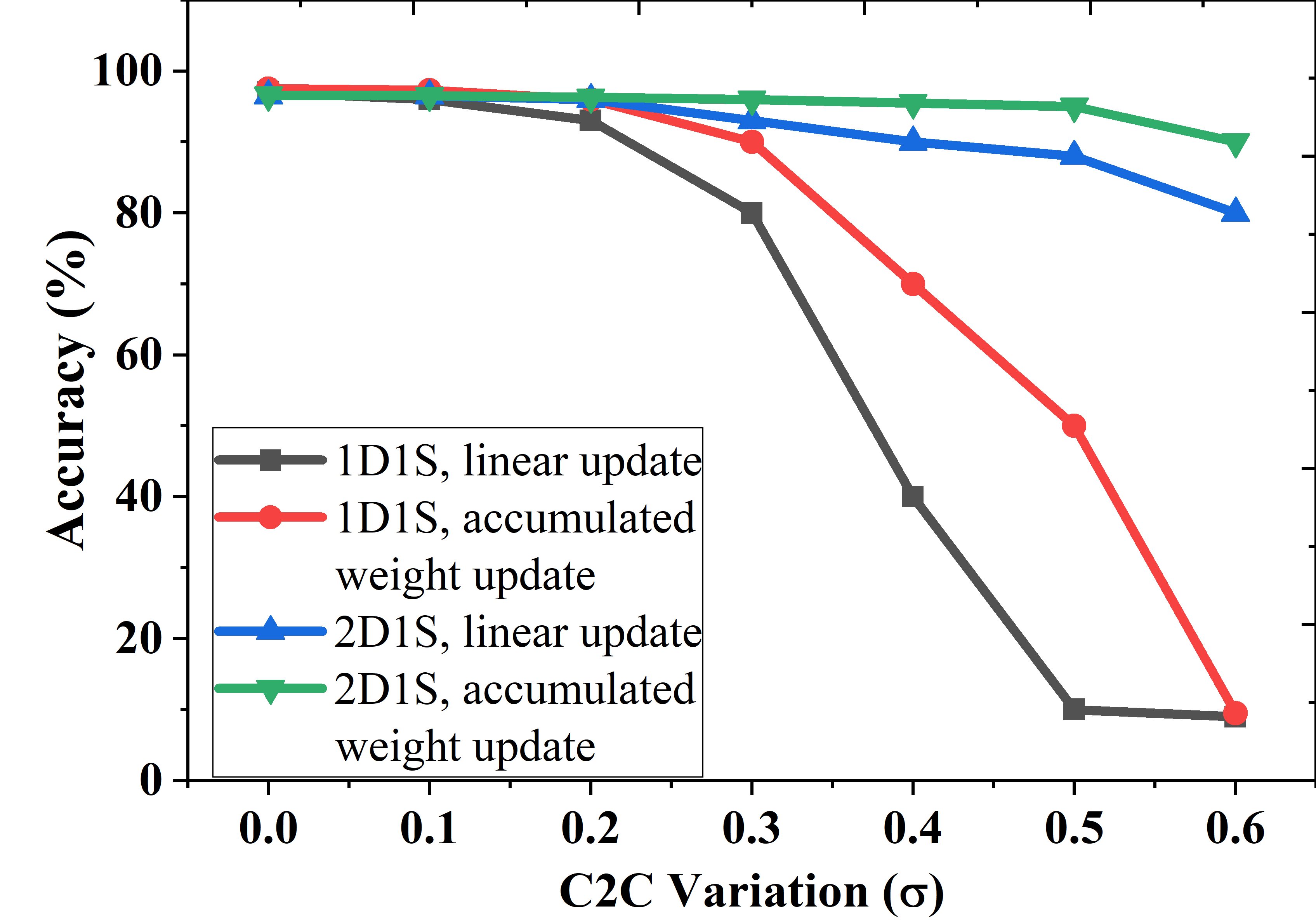}}
\caption{Effect of C2C variations on 1D1S and 2D1S synaptic architectures with and without the accumulated weight update.}
\label{C2C_on_1D1S_2D1S}
\end{figure}

In \cite{1D1S_Truong}, the authors demonstrated that the 1D1S architecture improves the areal density of crossbar arrays by twofold and reduces energy consumption by 48\% compared with the 2D1S architecture. In the CIMulator platform, we define a default reference conductance $G^{REF}=\frac{1}{2}(G_{max}-G_{min})$. However, in CIMulator, the 1D1S architecture has greater sensitivity to device variations and reduced accuracy compared with the 2D1S architecture owing to the reduction in effective bit number. Moreover, the possible weight values are reduced by half because the representable normalized weights are just in $[-0.5, 0.5]$. The results shown in Fig.~\ref{C2C_on_1D1S_2D1S} with 7-bit weight systems reveal priority of the 2D1S architecture over the other.

The implementation of $I_i=\sum_{j=1}^{n}V_jg_{j,i}$ facilitates the execution of the matrix multiplication $\mathbf{W}{\mathbf{x}}$, where $\mathbf{W}$ is the weight matrix and $\mathbf{x}$ is the input vector. The summed current $I_i$ is calculated in shift and add registers, sent to an ADC, and then sent to the batch normalization and activation function circuit before being transferred to the next layer.

\subsection{Bit Precision and Weight Quantization}
Weight values are represented by the conductance of synaptic cells, which is set (programmed) by discrete input pulses. They are quantized into 2$^b$ levels, where $b$ is the bit precision, before being written into the crossbar array. If $b$ is larger, weights can be represented more accurately. However, a higher bit precision requires more energy, furthermore, number of possible conductance levels are limited. CIMulator uses the accumulated weight update technique \cite{accumulated} and batch normalization in its first and second version, respectively, to achieve acceptable accuracy at a reduced bit precision. A simulation for using accumulated weight algorithm to minimize the effects of C2C variation is shown in Fig.~\ref{C2C_on_1D1S_2D1S}. In the study, the training system calculates and accumulates weight changes in backpropagation until the sum exceeds a predefined threshold, typically set as the average weight change, then updating occurs. However, it will not work if the sum is less than the threshold. The batch normalization would be a solution to overcome the drawback. Nevertheless, additional circuitry is required to perform batch normalization in hardware; this circuitry could be as simple as (digital) bit shifting circuitry combined with an ADC gain adjustment.

\subsection{Batch Normalization - Enabler for Training Low Bit-Width Networks}
\begin{figure}[ht]
\centerline{\includegraphics[width=0.8\linewidth]{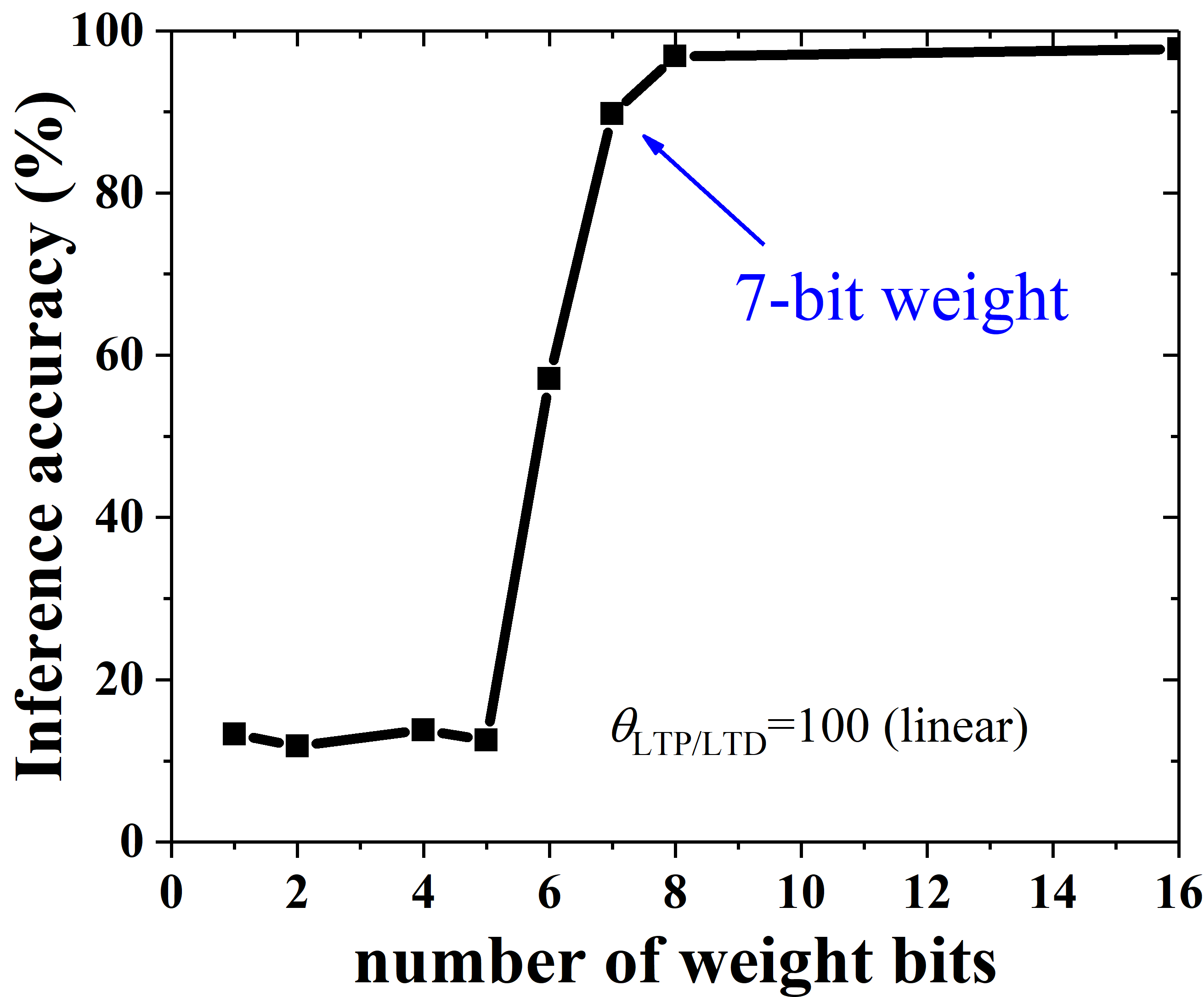}}
\caption{Simulation by the CIMulator platform indicates that with 7-bit weight and over, a neuromorphic accelerator system can achieve high training accuracy of greater than 90\% with conventional linear weight update algorithm}
\label{results_linear_update}
\end{figure}

In a neuromorphic accelerator system, the number of weight bits (or bit-width) strongly affects accuracy. A high-precision CIM system (with more weight bits) can achieve favorable training outcomes; however, it requires precise distinction between device conductance levels or a high ON/OFF ratio. Moreover, D2D and C2C variations have a greater effect on these systems. A 6-bit weight has been demonstrated to be optimal for a conventional neuromorphic accelerator \cite{Chang, NeuroSim1}. In addition, \cite{Quantized_Hubara} and \cite{BNN-S.M. Yu} have demonstrated that the precision of weights and features must be at least 6 bits because backpropagation passes small errors (gradients) from the output layer to the input layer; these errors cannot be effectively represented with a lower bit precision. This conclusion is consistent with the simulation results presented in Fig.~\ref{results_linear_update}; this simulation was performed using the CIMulator platform set for the linear case ($\theta_{LTP}=\theta_{LTD}=100$), without variation ($\sigma_{C2C}=\sigma_{D2D}=0$), and the conventional linear weight update method \cite{Neuro-inspired computing}. Software solutions for training modern deep learning networks typically require 32-bit [or single-precision (FP32) format] weights \cite{Micikevicius}. The CIM system can achieve high accuracy with a limited number of (less than 5) weight bits and greater nonlinearity owing to batch normalization.

\begin{figure}[ht]
\centerline{\includegraphics[width=1.0\linewidth]{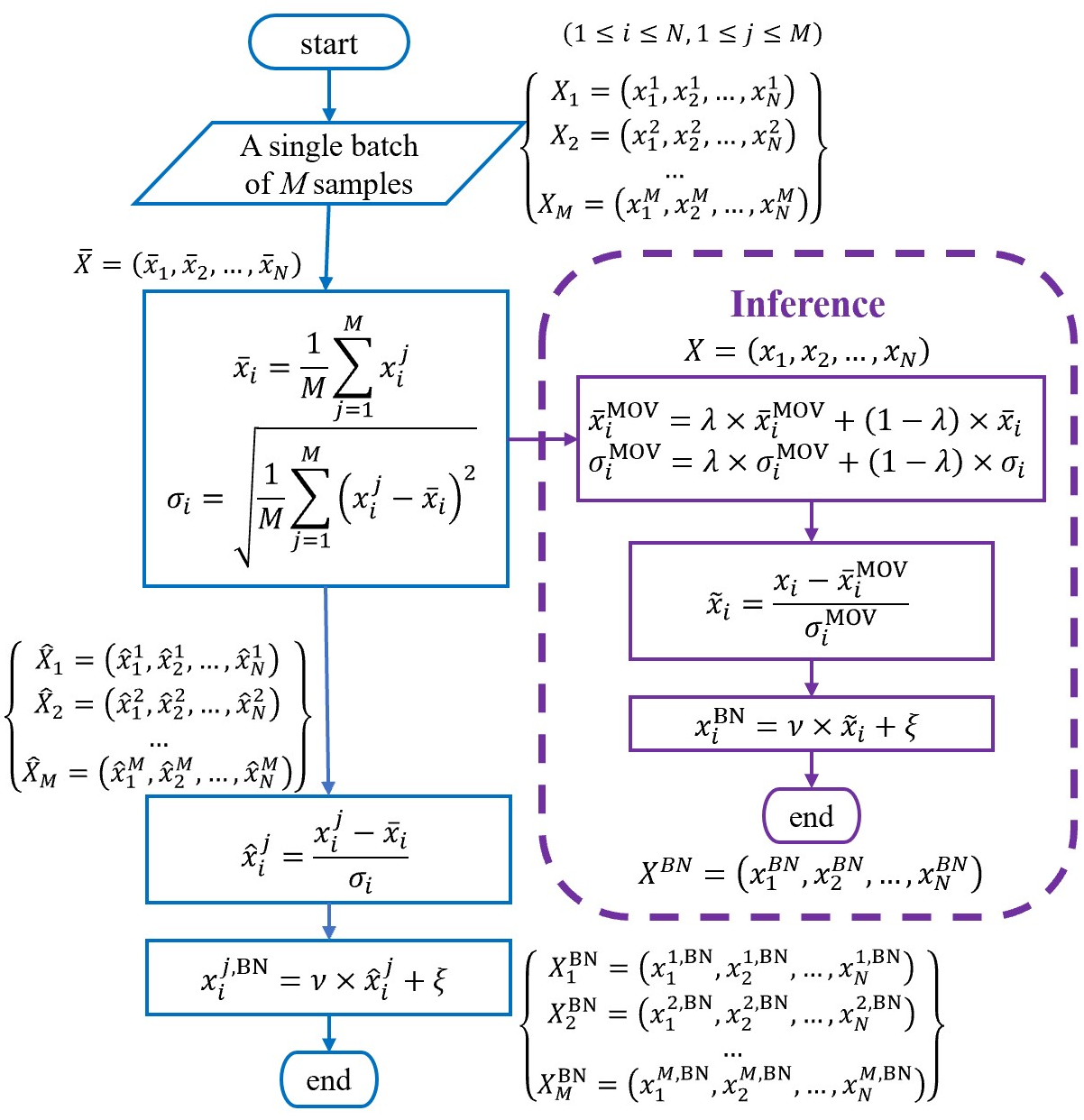}}
\caption{Batch normalization flowchart.}
\label{batchnorm}
\end{figure}

Batch normalization was initially proposed to mitigate the internal covariate shift problem \cite{BatchnormIoffe} during training. However, the technique can also enable deploying low-bit-weight deep learning systems efficiently in neuromorphic accelerators that avoid nonideal
problems of synaptic devices. 
Fig.~\ref{batchnorm} displays a flowchart of the batch normalization process which we apply in the CIMulator platform. Notice that, the ``learnable parameters'' $\nu$ and $\xi$ are learned during optimization, and $\lambda=0.1$ is a momentum scalar, which is used in inference only.

\section{Neuromorphic devices}

CIM neuromorphic accelerators perform the MAC operation within memory arrays comprising either volatile or nonvolatile synaptic devices \cite{ITRS_2013}. SRAM is volatile \cite{SRAM_Si}; RRAM \cite{RRAM01, RRAM02}, ferroelectric memory (FeRAM) \cite{FeRAM}, phase-change memory (PCM) \cite{PCM01, PCM02, PCM03}, magnetoresistive random access memory \cite{MRAM01, MRAM02}, and conventional embedded-flash \cite{Flash} are nonvolatile. In this paper, we primarily consider three common types of synaptic devices: RRAM, FeRAM 
, and SRAM 
. These are used as the basic weight-storing elements in the crossbar arrays for the MLP, CNN, and SNN architectures. In the platform, users may select the type of device for simulation by setting the appropriate input parameters prior to running the simulation.


\subsection{RRAM-Based Neuromorphic Accelerator}
In neuromorphic applications, RRAM has various advantages, such as simplicity of fabrication, low cost, high density, multilevel states, and flexibility in terns of material composition, structures, and behaviors. However, its disadvantages, which must be overcome at both the software and hardware levels, are its insufficient reliability, C2C variations due to random telegraphic noise, intrinsic D2D variability, and forming requirements \cite{NVM_review}. Moreover, RRAM devices often have nonlinearity, resulting in inaccurate weight writing and poorer accuracy rates than those of systems using linear devices.

RRAM is typically a metal\textendash insulator\textendash metal structure; the thin insulator is used as a switching material through the alteration of its resistance by using a voltage pulse to transition it between a high-resistive state (HRS) and low-resistive state (LRS). With a sufficiently low voltage, this switching may be partial, enabling the establishment of multiple resistance states. Relevant RRAM HRS/LRS statistics are modeled in \cite{CIMulator_v1}. The conduction mechanism depends on the material, but it typically involves the formation of conductive filaments that include oxygen vacancies. In neuromorphic systems, RRAM with tunable conductance is used as a synaptic element to store  neural network weights. In general, a compact and simple array structure forms a weight matrix. Each RRAM device is located at the cross point of a WL and a BL. These analog synaptic weights in real hardware have a limited conductance range. Therefore, we can normalize RRAM device conductance by using the following equation:

\begin{equation}
w=\frac{G\times(W_{max}-W_{min})}{G_{max}-G_{min}}, W_{min}\le{w}\le{W_{max}}
\label{w_RAM}
\end{equation}
where $G$, $G_{max}$, and $G_{min}$ are the actual conductance, maximum conductance, and minimum conductance of the RRAM device, respectively, and $W_{max}=1.0$ and $W_{min}=W_{max}\times{I_{ON/OFF}}$ are the defined maximum and minimum weights, respectively, $I_{ON/OFF}=\frac{G_{min}}{G_{max}}$ is called the ON/OFF ratio.

The conductance of a single RRAM device is set nonlinearly by a sequence of input voltage pulses. The following formulas can be used for  CIMulator:

\begin{equation}
w_{LTP}=\beta_{LTP}(1-e^{(-\frac{P}{\alpha_{LTP}})})+W_{min}
\label{LTP}
\end{equation}

\begin{equation}
w_{LTD}=-\beta_{LTD}(1-e^{(\frac{P-P_{max}}{\alpha_{LTD}})})+W_{max}
\label{LTD}
\end{equation}

\begin{equation}
\beta_{LTP/LTD}=\frac{W_{max}-W_{min}}{1-e^{(-\frac{P_{max}}{\alpha_{LTP/LTD}})}}
\label{beta}
\end{equation}

\begin{equation}
\alpha_{LTP/LTD}=\theta_{LTP/LTD}\times{P_{max}}
\label{alpha}
\end{equation}
where $\theta_{LTP/LTD}$ is the LTP or LTD curve nonlinearity factor, which can be determined by fitting the measured conductance to the voltage pulses; $w_{LTP}$ and $w_{LTD}$ are the LTP or LTD weight values, where $w_{min}$ and $w_{max}$ are the corresponding minimum and maximum values; and $P$ is the number of input pulses with maximum and minimum values of $P_{max}$ and $P_{min}$, respectively. Equations (\ref{LTP})\textendash(\ref{alpha}) are based on the derivations in \cite{NeuroSim2}, which demonstrated the nonlinear relationship between updated conductance and the number of pulses.

Weight updating (setting synaptic device conductance) is executed after each backpropagation process. For a 2D1S RRAM-based synaptic array, the update rule is presented as Algorithm 1. The notation $\nint{x}$ represents the round function, which returns the nearest integer value of $x$. The number of program pulses is $\nint{P^{LTP}_{j,i}}$, where $P^{LTP}_{j,i}$ is calculated as follows:

\begin{equation}
P^{LTP}_{j,i} = -\alpha_{LTP} \times \log\left(1- \frac{w^{NEW}_{j,i}-W_{min}}{\beta_{LTP}} \right)
\label{P_LTP}
\end{equation}
$\Delta w_{j,i}$ is the change in weight between nodes $i$ and $j$; $G^+_{max}=G^-_{max}=G_{max}$ and $G^+_{min}=G^-_{min}=G_{min}$ are maximum and minimum conductance values of positive and negative synaptic devices, respectively; and $W_{max}, W_{min}$ are the manually set maximum and minimum weights, respectively.


\SetKwComment{Comment}{//}{}
\RestyleAlgo{ruled}
\begin{algorithm} [hbt!]
\caption{Weight update algorithm}\label{alg:one}
\KwData{$\Delta w_{j,i} \neq 0$, $w^{OLD}_{j,i}$}
\KwResult{new $g^{+}_{j,i}$, new $g^{-}_{j,i}$}

\Comment{Evaluate new weight value:}
$w^{NEW}_{j,i} \gets w^{OLD}_{j,i} - \Delta w_{j,i}$ \;

\If{$g^{+}_{j,i} \neq G^+_{min}$} {
    $g^{+}_{j,i} \gets G^+_{min} $ \Comment*[r]{reset $g^{+}_{j,i}$}
    }

\If{$g^{-}_{j,i} \neq G^-_{min}$} {
    $g^{-}_{j,i} \gets G^-_{min}$ \Comment*[r]{reset $g^{-}_{j,i}$}
    }
\Comment{Program synaptic devices:}
\If{$w^{NEW}_{j,i} \neq 0$}{
    \Comment{Evaluate number of input pulses:}
    $P_{j,i} \gets \nint{P^{LTP}_{j,i}}$
    
    \Comment{Set value for $g^{+}_{j,i}$ or $g^{-}_{j,i}$:}
    \eIf{$w^{NEW}_{j,i} > 0$} {
        $g^{+}_{j,i} \gets \frac{G^+_{max}-G^+_{min}}{1-e^{-\frac{P_{max}}{\alpha_{LTP}}}} \left(1-e^{-\frac{P_{j,i}}{\alpha_{LTP}}}\right) + G^+_{min} $ 
        }{
        $g^{-}_{j,i} \gets \frac{G^-_{max}-G^-_{min}}{1-e^{-\frac{P_{max}}{\alpha_{LTP}}}} \left(1-e^{-\frac{P_{j,i}}{\alpha_{LTP}}}\right) + G^-_{min} $  
        }
    }
\end{algorithm}

The algorithm is described as follows. Consider $\Delta w_{j,i}$ be obtained from the backpropagation process. If $\Delta w_{j,i}=0$, no update is executed. Otherwise, the new weight value is estimated as $w^{NEW}_{j,i}=w^{OLD}_{j,i}-\epsilon\Delta w_{j,i}$, where $\epsilon$ is the learning rate. Both two RRAM devices in ``positive" and ``negative" synaptic arrays, which are representing for $w^{OLD}_{j,i}$, are reset (erased). The number of program pulses $P$ is calculated using (\ref{P_LTP}). If the sign of the new weight is positive ($w^{NEW}_{j,i}>0$), only $g^{+}_{j,i}$ is updated; otherwise, $g^{-}_{j,i}$ is set by $P$ program pulses. Notice that, this algorithm is effective for 2D1S architecture, and the weight values are stored in one of pair of RRAM devices. This process is more convenient than conventional weight update algorithms because the conductance is altered in only one device of the pair by applying positive input pulses based on the LTP curve function. However, the drawback of this algorithm is that the memory must be reset in every training epoch.

\subsection{FeFET-Based Neuromorphic Accelerator}
Recently, considerable efforts have been made to produce FeFETs that can be used as reliable elements for synaptic crossbar arrays. In our recent work \cite{FinFET_De, FeFET_Lu}, we demonstrated that FeFETs are superior to other emerging nonvolatile memory devices because of their fast switching, high ON/OFF ratio, excellent linearity, bidirectional operation, and favorable endurance. The cited study also included a helpful demonstration of the use of Hf$_{0.5}$Zr$_{0.5}$O$_2$ (HZO) FeFETs as synaptic cells for neuromorphic accelerator applications. In addition, unlike RRAM devices, the synaptic properties of FeFETs are highly dependent on $V_{gs}$. The compact model presented in \cite{FeFET_Lu} is useful for evaluating figures of merit, such as power consumption, circuit delay, and training accuracy, under the consideration of bias dependencies. Moreover, FeFETs are sufficiently compact for use in neuromorphic circuit design processes in platforms such as CIMulator. Other studies have also demonstrated the advantages of HZO-based FeFETs, including their fast WRITE speed \cite{FeFET_fast}, low-power programming, and high endurance. Another study \cite{FeFET_high_endurance} demonstrated that ferroelectric synaptic cells have excellent potential for use in CIM chips.

For FeFETs to be used in practice, they must achieve long-term retention. In \cite{Fe-FinFET}, we investigated the retention characteristics of Fe-FinFET devices with metal\textendash ferroelectric\textendash insulator\textendash semiconductor gate stacks. The study demonstrated that the accuracy of neural networks using these devices decreased over time because of the degradation of polarization and injection of charged carriers in the gate stack; the study also discussed strategies to mitigate this degradation.

In practical computing applications, circuits and devices must have a wide operating temperature range and must be applicable under conditions involving   temperature fluctuations that are caused either changes in ambient temperature in edge devices or by heat dissipation from chips. In \cite{FeFET_in_BNN} a CIMulator platform was used to simulate neural network training on an FeFET-based CIM accelerator at a single temperature (e.g., room temperature) and inference at another temperature. We demonstrated that an optimized binarized neural network can adequately withstand temperature fluctuations. Nevertheless, ferroelectricity disappears at the material limit given by the Curie temperature; thus, the operating temperature must be maintained below the Curie temperature with sufficient margin; otherwise, the weights would be erased.

The weight update process for the FeFET-based synaptic array is similar to that for an RRAM-based accelerator. However, the nonlinear relation between the number of program pulses and conductance is different from those in (\ref{LTP}) and (\ref{LTD}); it is dependent on which type of FeFET device is used.

\subsection{SRAM-Based Neuromorphic Accelerator}
CIM-oriented SRAM cells have more transistors than do regular 6T-SRAM cells. These additional transistors improve the READ stability by decoupling the WRITE and READ ports. A well-known SRAM-cell-based CIM architecture that uses 7T- and 8T-SRAM cells is presented in \cite{CIMAT_SRAM}.


In CIM, the MAC operation is realized by using  crossbar arrays of 7T- and 8T-SRAM cells. The MAC operation entails multiplying the weights of the neural network (SRAM cell data) by the pulse input applied along the read WLs (RWLs) of the SRAM cell. The total multiplied output can be derived by summing the currents from SRAM cells in the same column along the read BLs (RBLs) in accordance with the Kirchhoff current law (KCL) \cite{KCL}. Prior to the execution of the CIM operation, the RBL capacitors are precharged to $V_{dd}$; the voltage drop is proportional to the product of the multiplication of the data stored in the SRAM cell $W$ by the discharge current $I$ in accordance with a voltage pulse of finite duration $T_{WL}$ applied to the RWL, as presented in (\ref{V_SRAM}):

\begin{equation}
\Delta{V}=W\frac{1}{C_{BL}}\int_{0}^{T_{WL}}I(t) \,dt
\label{V_SRAM}
\end{equation}

\begin{figure}[ht]
\centerline{\includegraphics[width=1.0\linewidth]{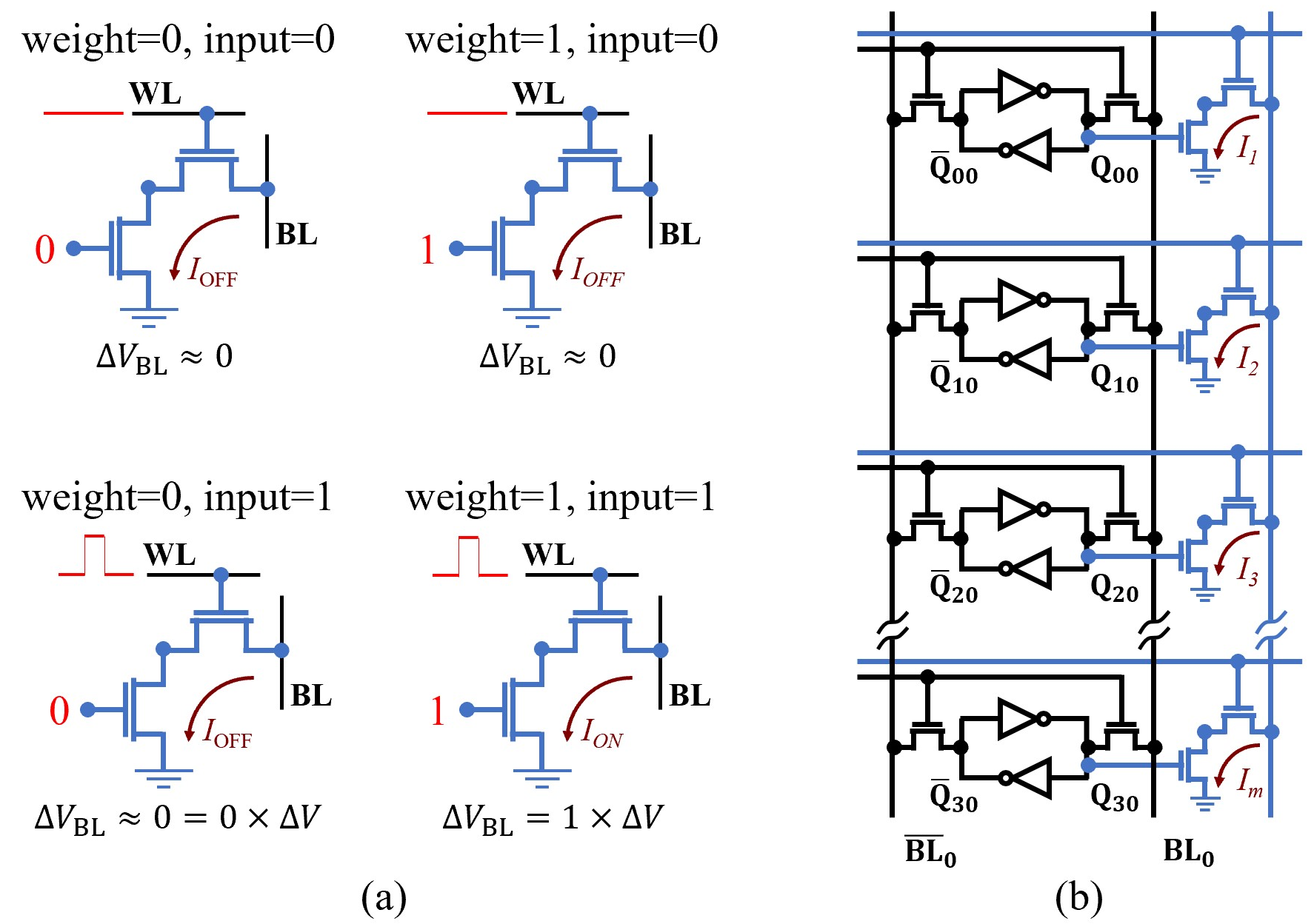}}
\caption{(a) Multiplication operation for various cases (input and SRAM weight being 1 or 0) by using SRAM cells. (b) MAC operation using KCL.}
\label{SRAM_MAC}
\end{figure}

Fig.~{\ref{SRAM_MAC}}(a) presents the multiplication results for four possible cases. The total discharge current along the RBLs can be derived using (\ref{VBL_SRAM})  and is presented in Fig.~{\ref{SRAM_MAC}}(b).

\begin{equation}
\Delta{V_{BL}}=\sum_{i=1}^{m}W_i\frac{1}{C_{BL}}\int_{0}^{T_{WL}}I_i(t) \,dt
\label{VBL_SRAM}
\end{equation}

The discharge current from the RBL capacitance is nonlinear; this is referred to as MAC nonlinearity. The MAC nonlinearity ($MAC_{NLOP}$) of a 4-bit flash ADC is presented in \cite{CFET_JEDS}. MAC nonlinearity can be modeled using the parameter $\theta_{SRAM}$, which is obtained by fitting the ADC output, as described in (\ref{alpha_SRAM}), (\ref{beta_SRAM}), and (\ref{MAC_SRAM}), in accordance with the ADC specifications.

\begin{equation}
\alpha_{SRAM}=\theta_{SRAM} \times 2^{ADC_{bits}-1}
\label{alpha_SRAM}
\end{equation}

\begin{equation}
\beta_{SRAM}=\frac{1-\frac{V_{min}}{V_{max}}}{1-e^{-\frac{2^{ADC_{bits}-1}}{\alpha_{SRAM}}}}
\label{beta_SRAM}
\end{equation}

\begin{equation}
MAC_{NLOP}=-\alpha_{SRAM} \times \ln \left( S \right)
\label{MAC_SRAM}
\end{equation}
where $S=\left|  \frac{V}{V_{max}}-\frac{V_{min}}{V_{max}}-\beta_{SRAM}\right|/\beta_{SRAM}$.

\section{Neural Network Training Examples}
This section presents neural network training simulations for various neuromorphic accelerator schemes using the CIMulator platform. We investigated CIM accelerators realized using RRAM, FeRAM, and SRAM. The platform supports a variety of datasets, including MNIST and CIFAR-10 datasets, as well as datasets for special applications, namely a WBC dataset. In addition, neural network customization can be performed to improve image recognition. The CIMulator simulations were implemented through a high-performance computer, the Tesla V100, using both CPUs and GPUs. The process is described in detail in the following subsections; the key results are summarized in Table \ref{Table_MLP_CNN_results}.

\subsection{CIMulator for MLP Model with the MNIST Dataset}

\subsubsection{MLP for MNIST dataset with RRAM-based synaptic cells}
\begin{figure}[ht]
\centerline{\includegraphics[width=1.0\linewidth]{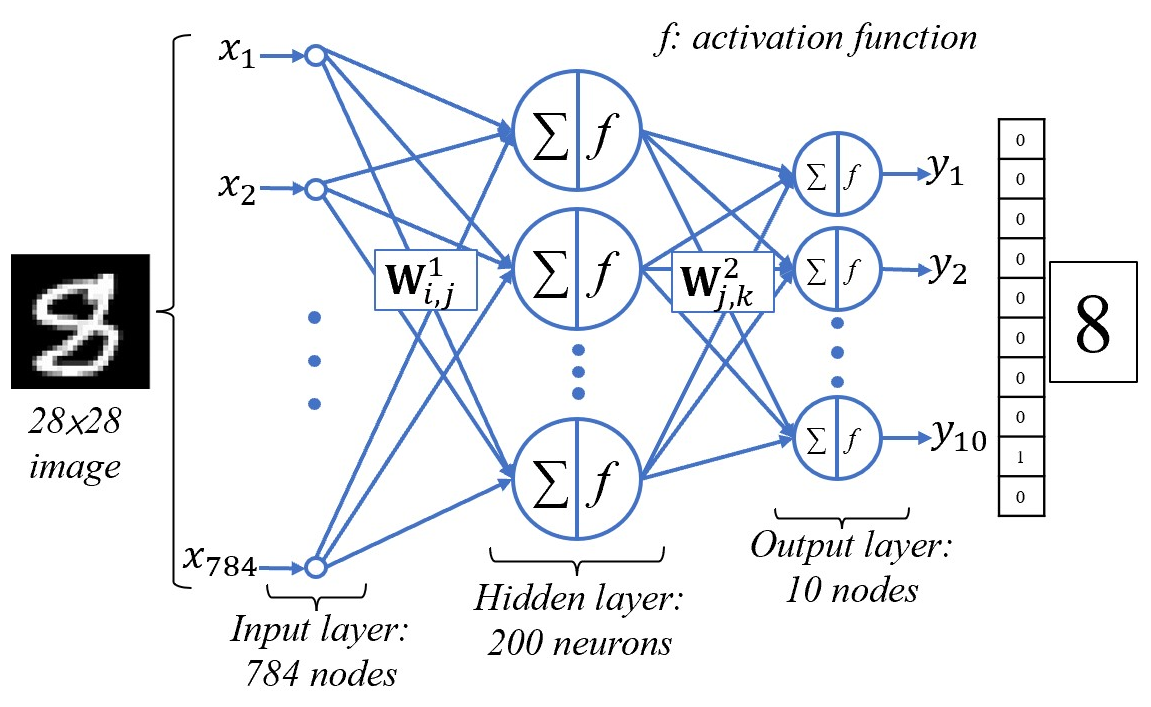}}
\caption{Conventional MLP architecture applied to the MNIST dataset in the CIMulator platform.}
\label{MLP}
\end{figure}

A conventional MLP architecture for the MNIST dataset \cite{MNIST} is illustrated in Fig.~\ref{MLP}. It includes 784, 200, and 10 nodes in the input, hidden, and output layers, respectively. The dataset comprises 60,000 training and 10,000 inference samples. We used the CIMulator platform to simulate the RRAM-based crossbar array in image recognition tasks. The device was made from Ni/HfO$_2$/TiN and was fabricated by our group \cite{CIMulator_v1}. 


The process involved calibrating the device LTP/LTD characteristics by using models developed for emerging nonvolatile memory devices such as FeFET, RRAM, and PCM. The developed models facilitate performing device-based neural network simulations by enabling the incorporation of nonideal device characteristics, such as variations, linearity, and the ON/OFF ratio.


\begin{figure}[ht]
\centerline{\includegraphics[width=0.8\linewidth]{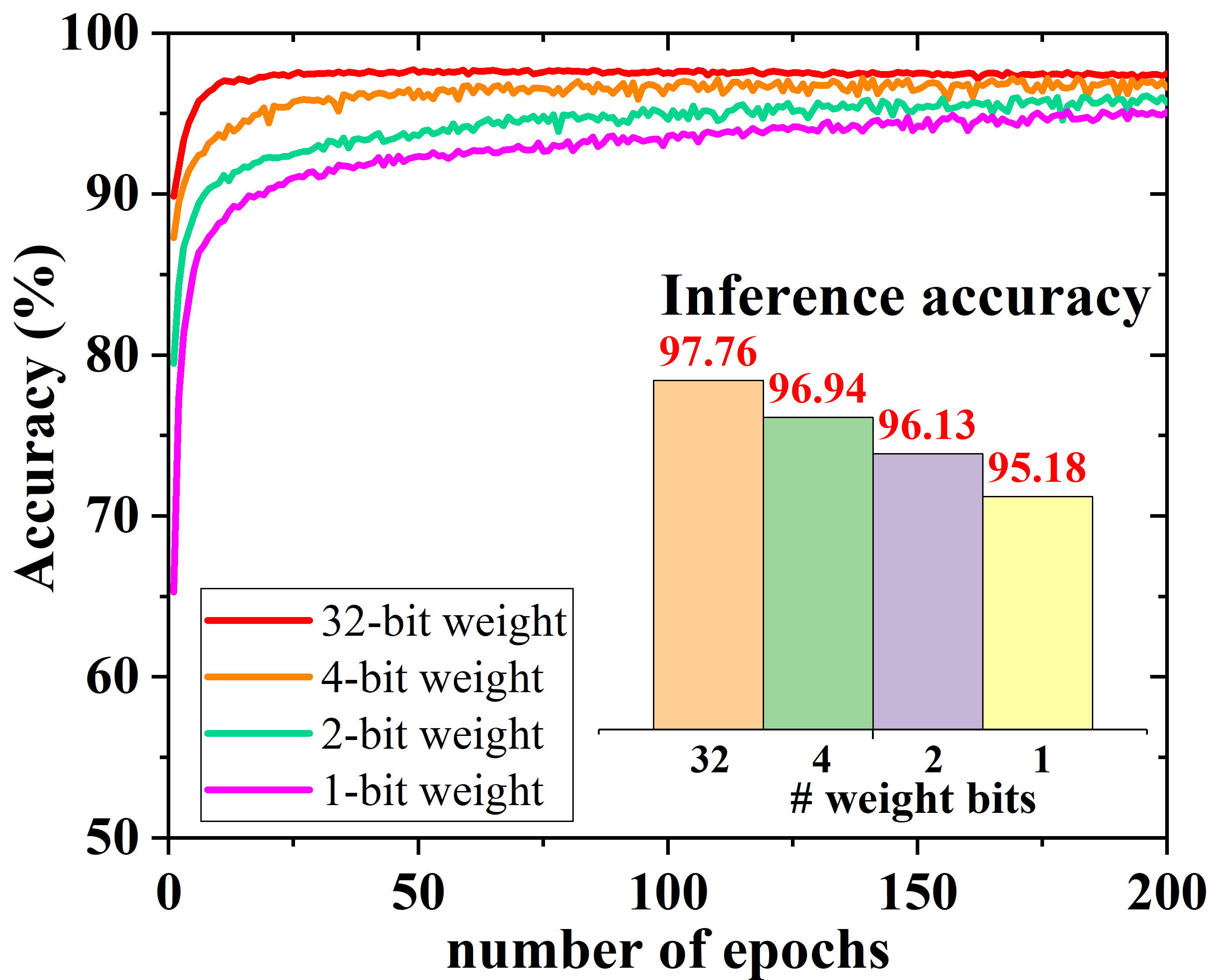}}
\caption{MLP accuracy derived for RRAM devices on the MNIST dataset under zero D2D variation.}
\label{MLP_MNIST}
\end{figure}

The test and inference accuracy are presented in Fig.~\ref{MLP_MNIST}. Higher bit weights engender higher accuracy with fewer number of epochs; lower bit weights require substantially more training to achieve high accuracy. The trained weights are saved for future use in inference tasks. Our simulation results revealed that the batch normalization technique enabled achieving approximately 96\% inference accuracy, even when very low bit widths (1 or 2 bits) were used. This value is higher than that obtained from NeuroSim (94.6\%)\ with a 6-bit weight system \cite{NeuroSim1} and only slightly lower than that of 32-bit weight systems.

\begin{figure}[ht]
\centerline{\includegraphics[width=0.8\linewidth]{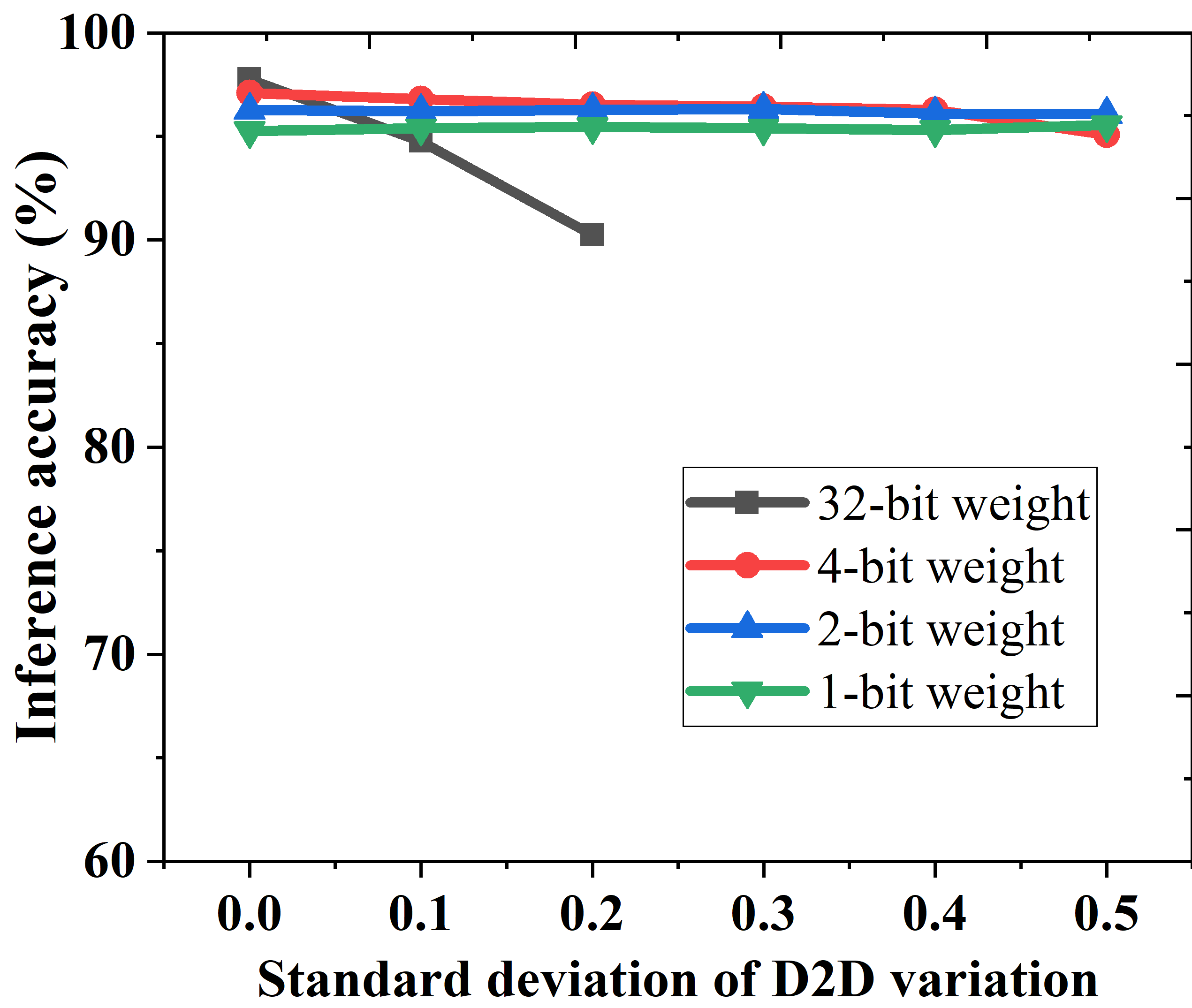}}
\caption{Inference accuracy of MLP on the MNIST dataset under various $\sigma_{D2D}$ values.}
\label{MLP_MNIST_D2D}
\end{figure}

We further investigated the effects of D2D variation on inference accuracy. The influence of D2D variation was greater for higher bit weight  than for lower bit weights. As indicated in Fig.~\ref{MLP_MNIST_D2D}, the inference accuracy decreased as the D2D variation increased when the bit weight was 32 bits; specifically, the accuracy decreased from 98\% with $\sigma_{D2D}=0.0$ to 95\% and 90\% for $\sigma_{D2D}$ of $0.1$ and $0.2$, respectively. A similar trend was observed when the bit weight was  4 bits. By contrast, the accuracy exhibited little variation with $\sigma_{D2D}$ when the bit weights were 2 and 1 bits, demonstrating an advantage of low-bit-width schemes.

The influence of nonlinearity on inference accuracy was also investigated using the LTP and LTD curves, which present the relation between number of positive or negative input pulses and the weight (or conductance) value. In the CIMulator platform, device nonlinearity was modeled using $\theta_{LTP/LTD,}$ whose value ranged from $0.01$ (extremely nonlinear) to $100$ (almost linear).

\begin{figure}[ht]
\centerline{\includegraphics[width=0.8\linewidth]{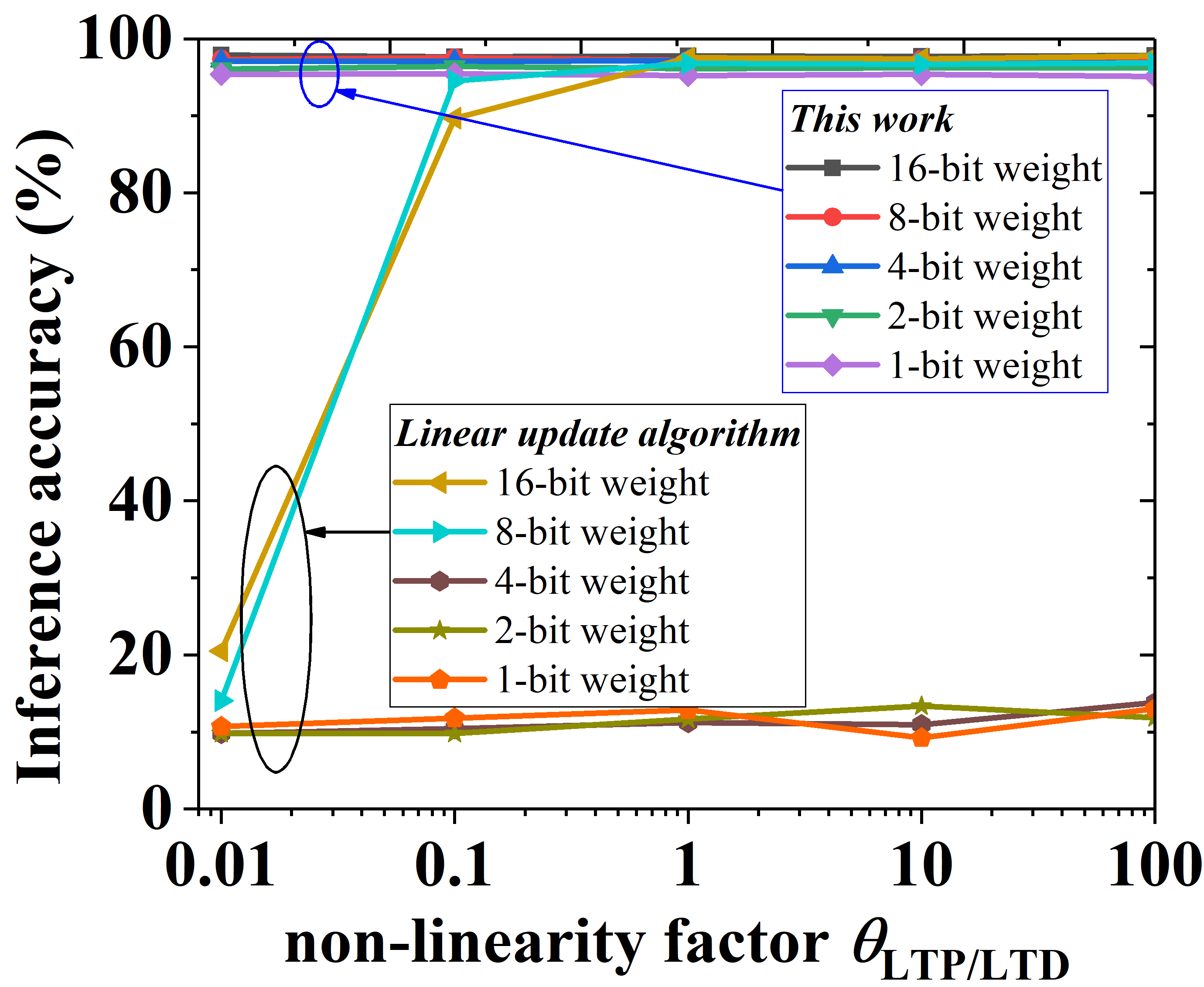}}
\caption{Inference accuracy of MLP on the MNIST dataset under various $\theta_{LTP/LTD}$ values.}
\label{MLP_MNIST_theta}
\end{figure}

In this investigation, we proposed a novel weight update algorithm in which all devices are reset before each weight update iteration. Thus, the weight value (represented by conductance) is written into devices using only the SET operation in accordance with the LTP formula (\ref{LTP}). Fig.~\ref{MLP_MNIST_theta} presents the effect of device nonlinearity on image recognition accuracy for single-bit and multibit weight architectures. Our simulations revealed that neural networks with this method were robust to device nonlinearity and achieved favorable training accuracy rates of \textgreater95\% with both low and high bit weights. By contrast, the linear weight update method (in which the relation between input pules and conductance is assumed to be linear) resulted in no learning when the bit weight was 1, 2, or 4 bits. Favorable accuracy could be achieved only when the bit weight was  8 or 16 bits and $\theta_{LTP/LTD}\geqslant0.1$.

\subsubsection{MLP for MNIST dataset using SRAM-based synaptic cells}
\begin{figure}[ht]
\centerline{\includegraphics[width=0.8\linewidth]{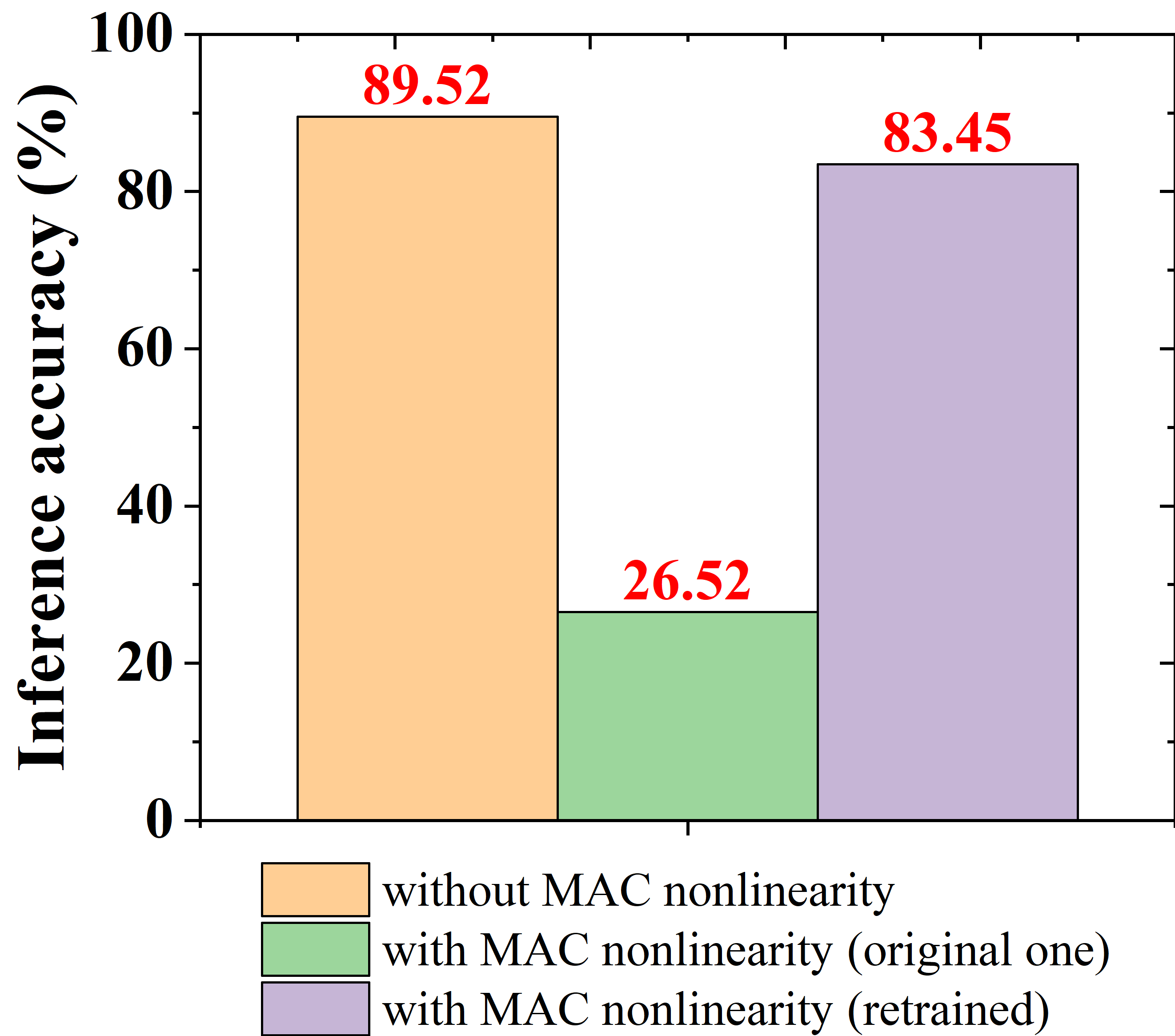}}
\caption{Simulation results for the FinFET-SRAM-based CIM device with 16 levels (4-bit weight).}
\label{SRAM_FinFET_ACC}
\end{figure}

For the image recognition task, the system was first trained with a pure software neural network to obtain the weights for the MLP model, which were then used in the inference phase with FinFET-based 8T-SRAM synaptic memory. Cases with and without MAC nonlinearity were simulated to observe the effect of MAC nonlinearity on the relationship between ADC inputs and outputs \cite{CFET_JEDS}. The effect of MAC nonlinearity on neural network accuracy was small; because SRAM cells have infinite endurance, the neural network can be retrained in few epochs to restore most of the lost accuracy. As displayed in Fig.~{\ref{SRAM_FinFET_ACC},} a recognition rate of 89.52\% was achieved in the case without MAC nonlinearity; introducing MAC nonlinearity reduced the accuracy to 26.52\%, but retraining restored the accuracy to 83.45\%.

\subsection{CIMulator Platform For CNNs}
In addition to MLP, we implemented CNNs such as LeNet-5, VGG-16, and a novel CNN architecture C4W-1 in CIMulator. C4W-1 was implemented for only the WBCs dataset; the digit $1$ indicates that CFW-1 is the first version of the network. Block diagrams of the CNNs are presented in the following sub-sections. A CONV block represents a series of convolutional layers, activation functions, a batch normalization layer, a dropout layer, a max pooling layer, and a fully connected layer. 



\subsubsection{LeNet-5 for MNIST dataset using Fe-FinFET-based synaptic cells}
\begin{figure}[ht]
\centerline{\includegraphics[width=0.8\linewidth]{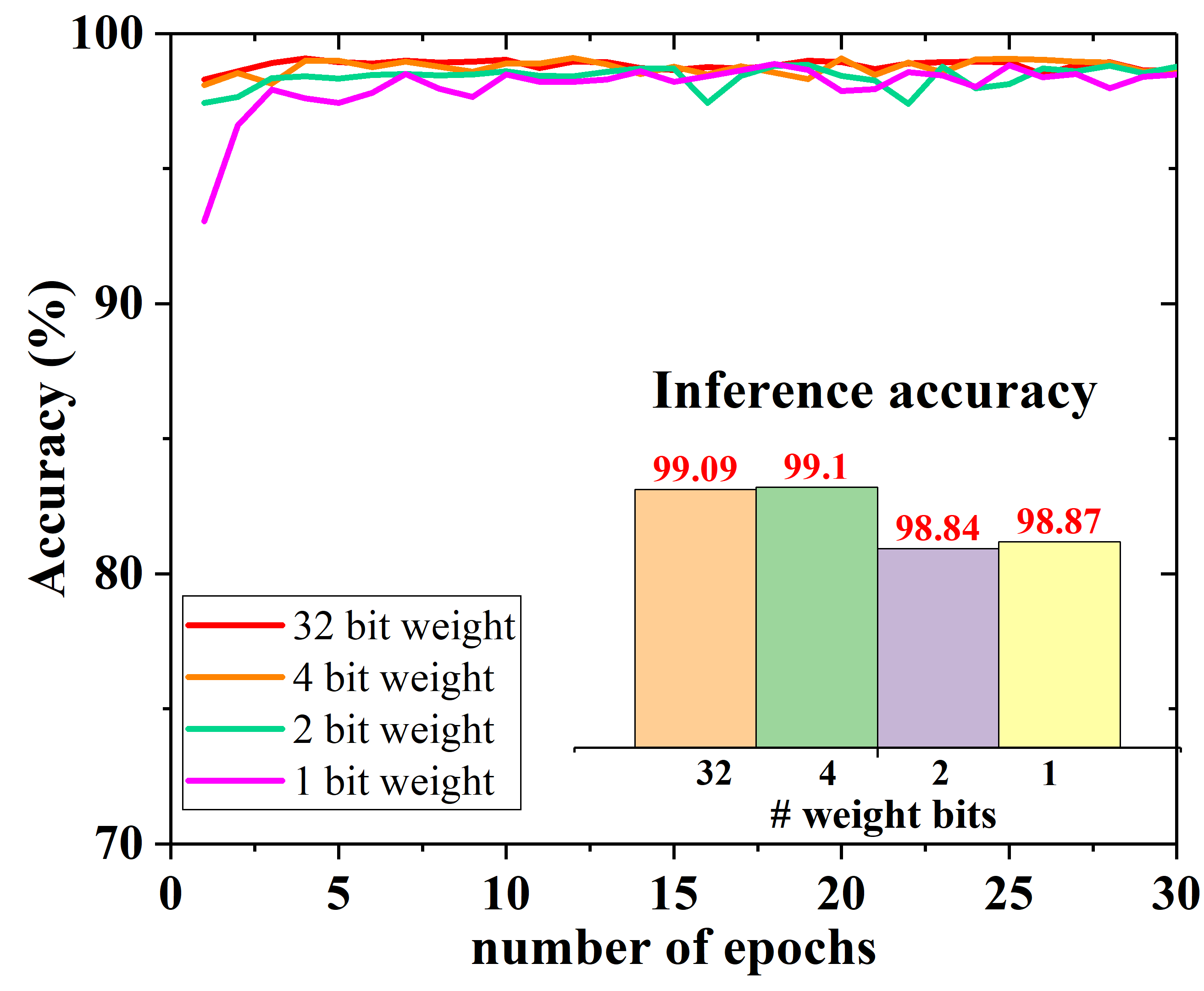}}
\caption{Accuracy of the LeNet-5 network architecture for the MNIST dataset with Fe-FinFET synaptic devices.}
\label{LeNet_MNIST}
\end{figure}

This section presents an example of the LeNet-5 network architecture for the MNIST dataset. LeNet-5 typically achieves higher accuracy than MLP does. Both low-bit-width and 32-bit (floating point) weight systems were investigated in our simulation. In the simulation, the synaptic device was a 10-nm Fe-FinFET proposed in \cite{FeFET_in_BNN}. The high- and low-channel conductances of this device were $3.05 \times 10^{-06}$ and $1.19  \times 10^{-08}$ S, respectively, at 237{ K} and a read voltage of 1.2 {V}. The ON/OFF ratio could reach 300. The online training and inference accuracy are presented in Fig.~\ref{LeNet_MNIST}. The accuracy of the 4-bit weighted synapses was more stable than that of the 1- and 2-bit weighted synapses and did not differ substantially from that of the 32-bit weight benchmark case. These results were similar to those obtained for the MLP simulation on the MNIST dataset.

\subsubsection{VGG-16 for Cifar-10 dataset using FeFET-based synaptic cells}
\begin{figure}[ht]
\centerline{\includegraphics[width=1.0\linewidth]{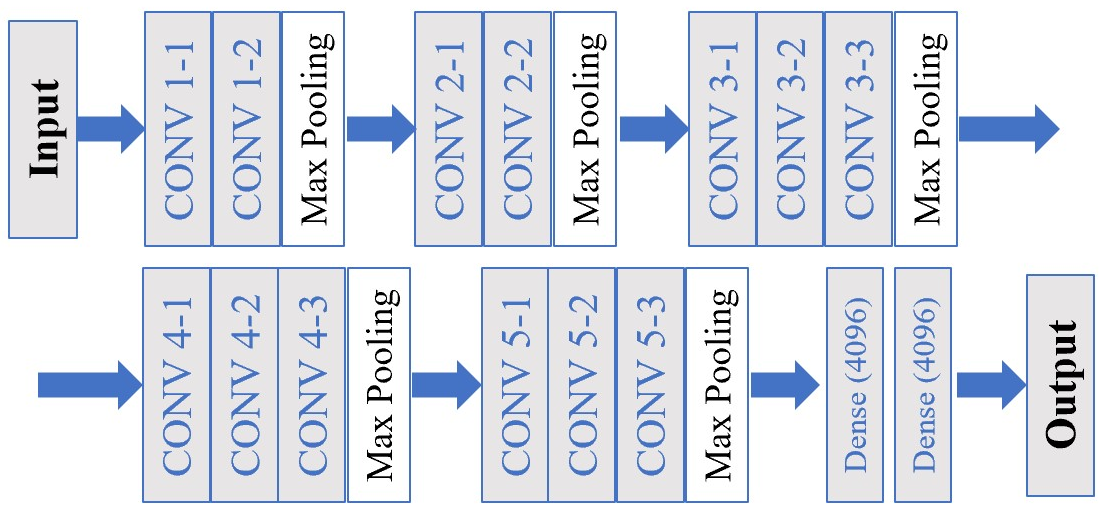}}
\caption{VGG-16 architecture used in the CIMulator platform.}
\label{VGG16_arc}
\end{figure}

\begin{figure}[ht]
\centerline{\includegraphics[width=0.8\linewidth]{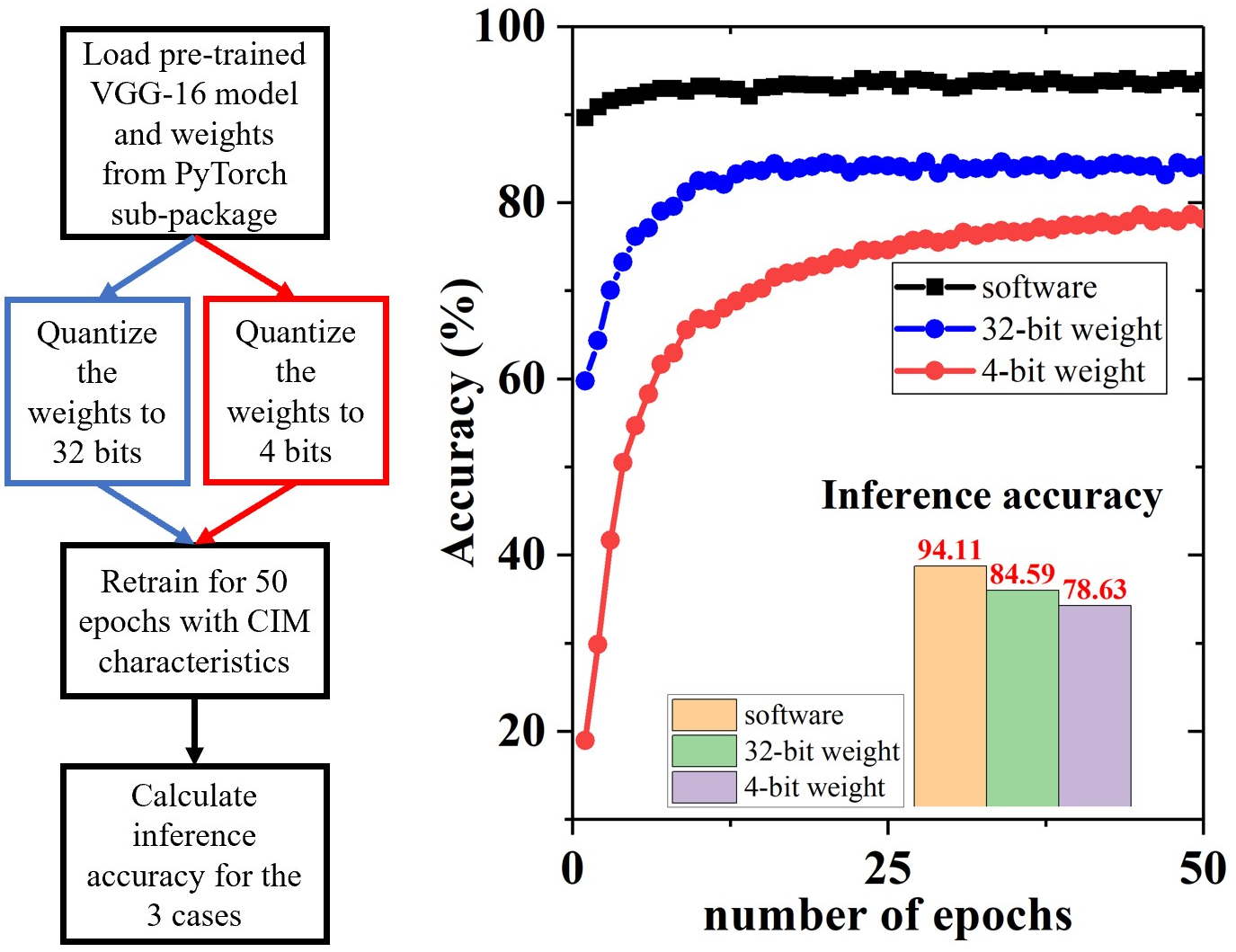}}
\caption{Accuracy for 32-bit and 4-bit weight VGG-16 on the Cifar-10 dataset with FeFET based synapses. D2D variability and nonlinearity were considered for the 4- and 32-bit weight but not software cases.}
\label{results_VGG_Cifar}
\end{figure}

VGG-16 \cite{VGG-16} is a commonly used deep CNN architecture that can be applied for image recognition on ImageNet \cite{ImageNet}, a large dataset of over 14 million images belonging to 1000 classes. We conducted a simulation by applying VGG-16 to the smaller dataset Cifar-10 \cite{Cifar10} comprising 60,000 $32 \times 32$ color images in 10 classes, with 50,000 images serving as training data and 10,000 images serving as test data. The HZO-based FeFET proposed in \cite{FeFET_Lu} was used in the accelerator. The CIMulator platform can not only enable the end-to-end execution of training, test, and inference processes but also enable the reuse of pretrained models derived from the platform or from available PyTorch frameworks for retraining and one-shot inference tasks. The software-pretrained VGG-16 weights and model from the PyTorch \textit{torchvision.models} subpackage \cite{PyTorch_pretrained_model} were first loaded; retraining was then performed for 50 epochs to obtain higher learning accuracy. The obtained models and weights were written into the crossbar array, and a realistic performance measurement of the emerging nonvolatile memory device crossbar was obtained by incorporating nonideal device characteristics, such as D2D and C2C variations, nonlinearity, and low-bit weights. An inference process was then performed to obtain the final accuracy results. The VGG-16 network architecture is presented in the Fig.~{\ref{VGG16_arc}}. For simplicity, each light blue convolutional block CONV represents a set of convolutional layers, batch normalization layers, and activation subblocks. The inference accuracy reached 84.6\% and 78.6\% for the 32- and 4-bit weights, respectively. As illustrated in Fig.~{\ref{results_VGG_Cifar},} the software solution accuracy (obtained from the PyTorch framework) was \textgreater90\%, which decreased to approximately 80\% after quantization to 4- and 32-bit weights and retraining to overcome device nonidealities.

\subsubsection{C4W-1 network for white blood cells dataset using RRAM-based synaptic cells}
\begin{figure}[ht]
\centerline{\includegraphics[width=1.0\linewidth]{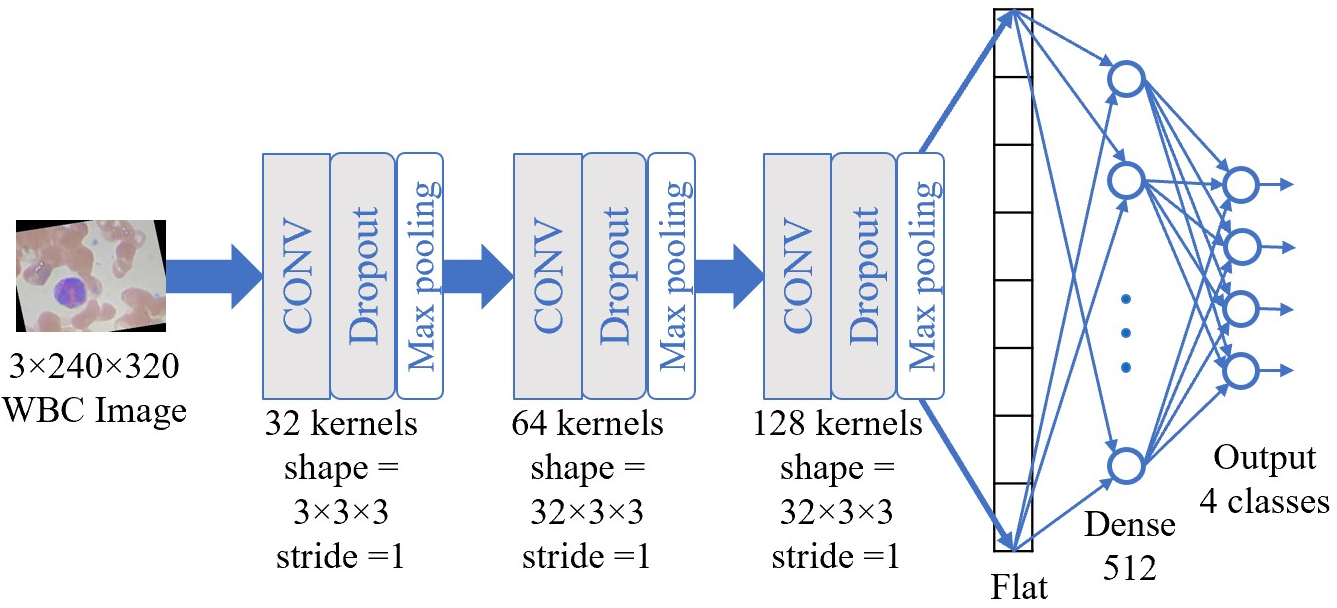}}
\caption{C4W-1 network architecture.}
\label{customCNN}
\end{figure}

\begin{figure}[ht]
\centerline{\includegraphics[width=0.8\linewidth]{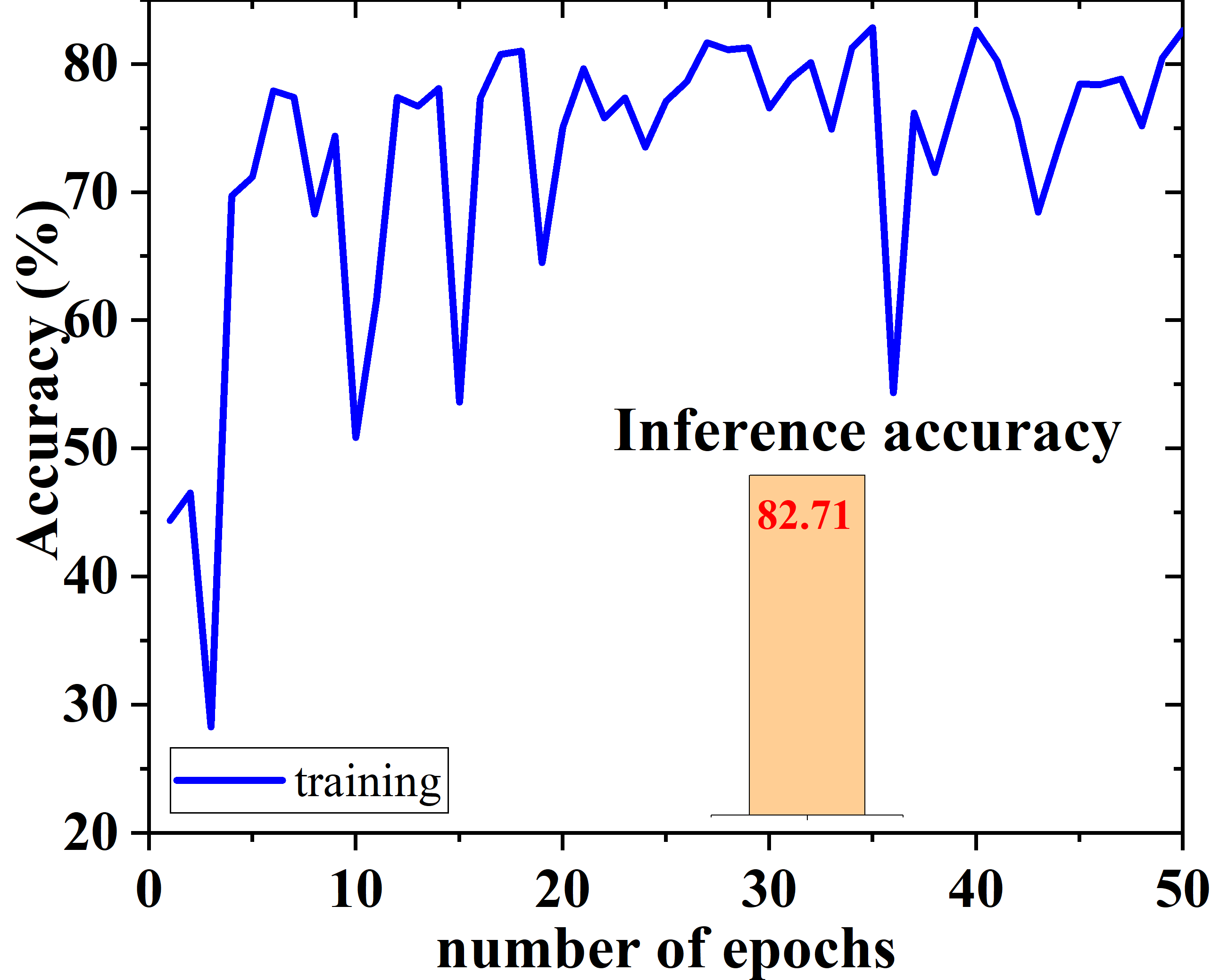}}
\caption{Accuracy rates of the 4-bit weight C4W-1 on the WBC dataset.}
\label{customCNN_WBC}
\end{figure}

The CIMulator platform can be applied flexibly to design and evaluate the performance of custom neural network architectures incorporating various synaptic devices. This section describes our simulation of a custom C4W-1 neural network for a WBC dataset \cite{WBC}. The configuration of this network is illustrated in Fig.~{\ref{customCNN}}; the key components are three convolution combinations, batch normalization, activation and dropout layers, and one dense layer. The network provided four outputs for the classification of four WBC types, namely eosinophil, lymphocyte, monocyte, and neutrophil. The dataset comprises 9957 training and 2487 test images. As displayed in Fig.~{\ref{customCNN_WBC},} the network achieved a high recognition accuracy rate (\textgreater80\%) when the optimizer was RMSProp, batch size was 60, number of training epochs was 50, and bit weight was only 4 bits. The neural network was realized using RRAM devices composed of Ni/HfO$_2$/TiN \cite{CIMulator_v1}.


These simulation results demonstrate that CNNs with low-bit-width weights can achieve high accuracy even on poor-quality biological datasets comprising unclear components along with components of interest. This results demonstrate that these complex image classification tasks can be deployed practically on neuromorphic accelerator hardware.

\begin{table*}[h]
\centering
\caption{Simulation parameters and results for MLP and CNN architectures on various datasets}
\begin{tabular}{lcccc}
\toprule
\textbf{}               & \textbf{MLP}  & \textbf{LeNet-5} & \textbf{VGG-16} & \textbf{Custom CNN} \\
\midrule
datasets               & MNIST         & MNIST            & Cifar-10        & WBC    \\
Type of device          & RRAM          & Fe-FinFET        & FeFET            & RRAM  \\
Material of device      & Ni/HfO$_2$/TiN \cite{CIMulator_v1} & 10nm HZO-based Fe-FinFET \cite{FeFET_in_BNN}    & HZO-based FeFET \cite{FeFET_Lu}   & Ni/HfO$_2$/TiN \cite{CIMulator_v1}      \\
Max. conductance (S)    & 1.04E-06      & 3.05E-06      & 0.68E-5      & 1.04E-06        \\
Min. conductance (S)    & 5.60E-08      & 1.19E-08      & 0.46E-10      & 5.60E-08        \\
Weight precision        & 1-bit weight & 1-bit weight    & 4-bit weight    & 4-bit weight        \\
Linearity ($\theta_{LTP}$, $\theta_{LTD}$)   & (0.2476, 0.2476)  & (9.2339, 0.4385)   & (0.67, 1,13)     & (0.2476, 0.2476)\\
D2D variation ($\mu, \sigma$)           & (0.0, 0.0)    & (0.0, 0.02)  & (0.0, 0.01)    & (0.0, 0.01) \\
C2C variation ($\mu, \sigma$)           & (0.0, 0.0)    & (0.0, 0.01)  & (0.0, 0.01)    & (0.0, 0.01) \\
GD algorithm            & SGD           & SGD             & Adam            & RMSProp \\
Learning rate           & 0.005         & 0.01            & 0.001           & 0.01    \\
Number of epochs        & 200           & 100             & 50               & 50      \\
On/Off-line training    & online        & online          & inference       & offline \\
Synaptic architecture   & 2D1S          & 2D1S            & 2D1S            & 2D1S    \\
Inference accuracy      & 95.18\%       & 99.03\%         & 78.63\%            & 80\%    \\
Running time per epoch (s)       & 9.5          & 41            &577               &--       \\
\bottomrule
\label{Table_MLP_CNN_results}
\end{tabular}
\end{table*}

\subsection{CIMulator Platform For Fully Connected SNN}
Inspired by biological information processing, SNNs have recently been proposed and developed as third-generation neural networks. In contrast to MLP and CNNs that perform vector matrix multiplication and use activation functions such as ReLU or Sigmoid functions, SNNs use spiking neurons as computational units; in such neurons, sparse and asynchronous signals are encoded as discrete neuron emission spikes. Our CIMulator platform supports unsupervised SNN training in which features are extracted using a biology-based learning rule known as spike-timing-dependent plasticity \cite{STDP}. In this study, SNNs realized using RRAM devices composed of TiN/HfO$_2$/TiN, Pt/HfO$_2$/TiON/TiN, or Ag/SrTiO$_3$/RGO/FTO were simulated using the CIMulator platform and an event driven algorithm \cite{Thesis_ITing}.

\begin{figure}[h]
\centerline{\includegraphics[width=1.0\linewidth]{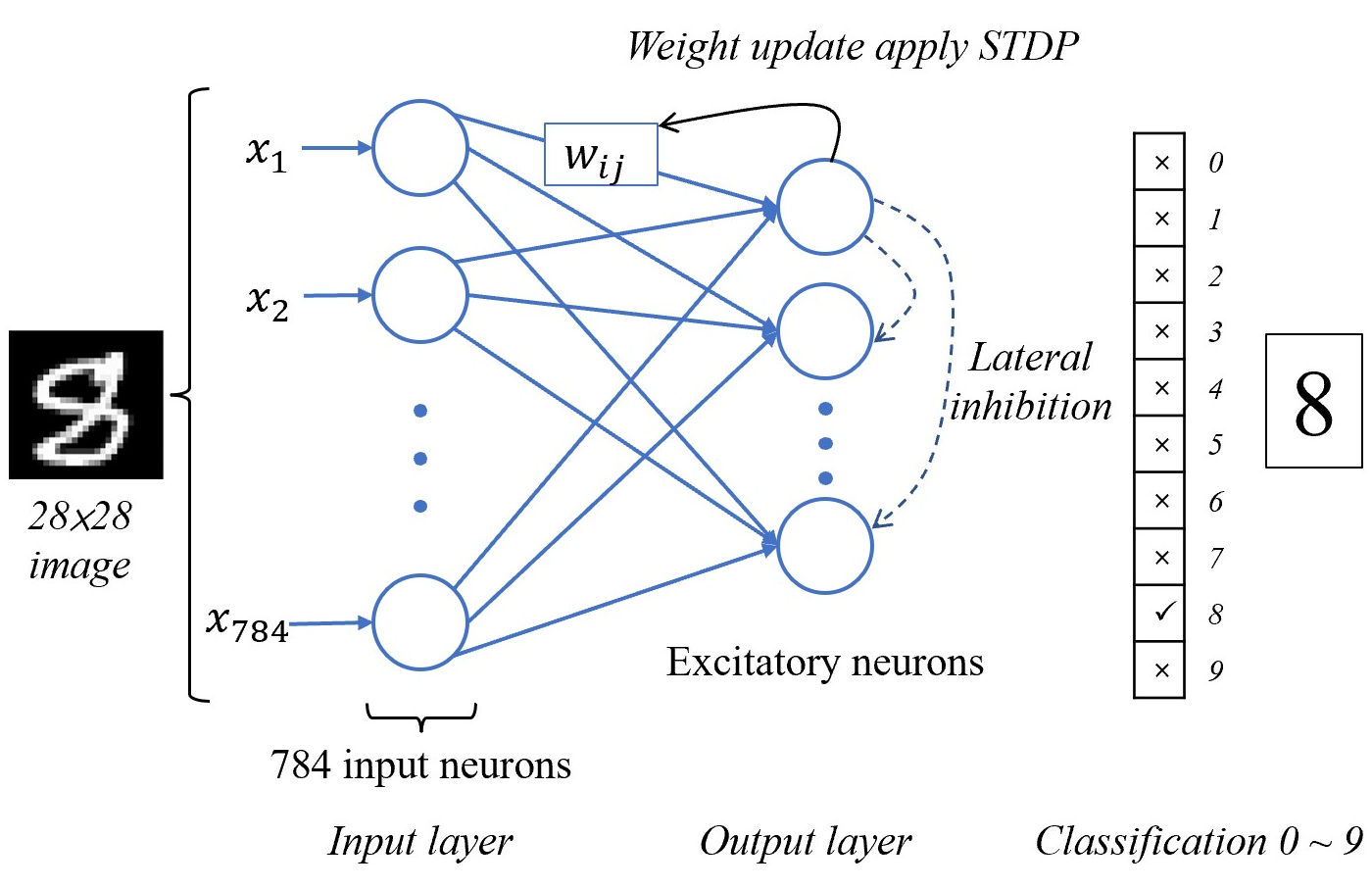}}
\caption{SNNs in CIMulator with 784 input and 400 output neurons.}
\label{SNN_architecture}
\end{figure}

Table \ref{Table_SNN} presents the aforementioned device and pulse characteristics. The measured characteristics were modeled using the fitting parameters ($A_+$, $A_-$, $\tau_+$, and $\tau_-$), and an SNN with 784 input and 400 output neurons achieved a maximum test accuracy of 88.81\% on the MNIST dataset.

\begin{table*}[h]
\centering
\caption{Simulation parameters and results for SNNs with RRAM-based CIM devices of various materials}
\begin{tabular}{lcccc}
\toprule
\multicolumn{2}{l}{\textbf{}} & \textbf{\begin{tabular}[c]{@{}c@{}}TiN/HfO$_2$/TiN \cite{SNN_TiN}\end{tabular}} & \textbf{\begin{tabular}[c]{@{}c@{}}Pt/HfO$_2$/TiON/TiN \cite{SNN_Pt}\end{tabular}} & \textbf{\begin{tabular}[c]{@{}c@{}}Ag/SrTiO$_3$/RGO/FTO \cite{SNN_Ag}\end{tabular}} \\ 
\midrule
\multirow{2}{*}{\begin{tabular}[l]{@{}l@{}}Weight ($\Delta G/G$)\end{tabular}} & max  & 0.96   & 0.78  & 0.60 \\ 
                                                                               & min  & 0.012  & 0.01 & 0.00074 \\ 
\multicolumn{2}{l}{\begin{tabular}[l]{@{}l@{}}Dynamic range \\($\Delta W_{max}/\Delta W_{min}$)\end{tabular}} & 80    & 78  & 810 \\ 
\multicolumn{2}{l}{On/Off ratio} & 25   & 190  & 1.5\\ 
\multirow{2}{*}{\begin{tabular}[l]{@{}l@{}}Voltage ($V$)\end{tabular}} & Set  & 0.5   & -1.0  & 1\\ 
                                                                       & Reset & -0.4   & 0.7   & -1\\ 
\multirow{2}{*}{\begin{tabular}[l]{@{}l@{}}Pulse scheme\end{tabular}}  & Pre   & 1.2$V$, 100$\mu s$  & 0.6$V$, 1$ms$ & 1$V$, 10$ms$\\ 
                                                                       & Post  & -2$V$, 100$\mu s$   & 0.6$V$, 1$ms$  & -1$V$, 10$ms$\\ 
\multirow{2}{*}{\begin{tabular}[l]{@{}l@{}}Fitting parameter\end{tabular}}  & $A_+$, $A_-$        & 0.96, -0.89    & 0.78, -0.44  & 0.60, -0.67\\ 
                                                                            & $\tau_+$, $\tau_-$  & 11$ms$, 66$ms$   & 32$\mu s$, 33$\mu s$   & 32$ms$, 31$ms$\\ 
\multirow{2}{*}{\begin{tabular}[l]{@{}l@{}}Accuracy (\%)\end{tabular}}      & Training  & 88.05   & 88.28   & 86.76 \\ 
                                                                            & Test      & 88.55   & 88.81   & 88.45 \\ 
\multirow{2}{*}{\begin{tabular}[l]{@{}l@{}}Running time\\($mins$)\end{tabular}}    & Training  & 43   & 44   & 45 \\
                                                                                        & Test      & 15   & 29   & 30 \\ 
\bottomrule
\label{Table_SNN}
\end{tabular}
\end{table*}

\section{Conclusion}
The CIMulator platform can simulate and evaluate the performance of neuromorphic accelerators with both conventional DNNs and SNNs. The platform supports various devices, including RRAM, FeRAM, PCM, and SRAM devices, and various neural network architectures. Realistic inference comprising spatial and temporal variation can be achieved by incorporating statistics obtained from device measurements. Techniques such as batch normalization can be used to model nonideal device characteristics, such as D2D variation, C2C variation, and low ON/OFF ratios. The obtained results reveal that high test and inference accuracy rates could be achieved even when low-bit-width weights were used. In the future, CIMulator will be upgraded to support more complex artificial neural network architectures and to provide latency, chip area, and energy consumption estimates. Moreover, SNN can be simulated in a consistent framework to enable direct comparison with the outcome of deep neural networks.



\section*{Acknowledgment}
This work was jointly supported by the Ministry of Science and Technology (Taiwan) grant MOST-110-2221-E-006-086-MY3, MOST-108-2634-F-006-008, MOST-109-2628-E-006-010-MY3 and is part of research work by MOST's AI Biomedical Research Center. We thank the National Center for High-performance Computing (NCHC) for providing computational and storage resources.


\end{document}